\documentclass{article}

\usepackage{PRIMEarxiv}

 \usepackage[utf8]{inputenc} 
\usepackage[T1]{fontenc}    
\usepackage[pdftex]{hyperref}       
\usepackage{url}            
 \usepackage{booktabs}       
 \usepackage{amsfonts}       
 \usepackage{nicefrac}       
 \usepackage{microtype}      
 \usepackage{lipsum}
 \usepackage{fancyhdr}       
 \usepackage{graphicx}       
 \graphicspath{{media/}}     

\title{Normalization of Lithuanian Text Using Regular Expressions
\thanks{\textit{\underline{Citation}}: 
\textbf{P.Kasparaitis. Normalization of Lithuanian Text Using Regular Expressions. Pages 1-21. DOI:2312.17660.}} 
}

\author{
  Pijus Kasparaitis \\
  Institute of Informatics, Faculty of Mathematics and Informatics \\
  Vilnius University \\
  Lithuania \\
  \texttt{pijus.kasparaitis@mif.vu.lt} \\
}

\begin{document}
\maketitle

\begin{abstract}
Text Normalization is an integral part of any text-to-speech synthesis system. In a natural language text, there are elements such as numbers, dates, abbreviations, etc. that belong to other semiotic classes. They are called non-standard words (NSW) and need to be expanded into ordinary words. For this purpose, it is necessary to identify the semiotic class of each NSW. The taxonomy of semiotic classes adapted to the Lithuanian language is presented in the work. Sets of rules are created for detecting and expanding NSWs based on regular expressions. Experiments with three completely different data sets were performed and the accuracy was assessed. Causes of errors are explained and recommendations are given for the development of text normalization rules.
\end{abstract}

\keywords{text-to-speech synthesis \and text normalization \and regular expressions \and Lithuanian}

\section{Introduction}
\label{s1}

Text-to-speech synthesis can be divided into two main stages: Natural Language Processing (NLP) and Digital Signal Processing (DSP) \cite{Dutoit97}. The component of NLP can be further decomposed into Text Normalization (sometimes denoted as Pre-processing), Morphological Analyzer, Contextual Analyzer, Syntactical Parser, Letter-to-Sound Rules and Prosody Generator. Usually, Text Normalization is the first stage of text-to-speech synthesis, while some authors, e.g., \cite{Huang01}, put Document Structure Detection stage before it. The present work deals with Text Normalization.

The text-to-speech synthesis system should be able to read out loud any given text. Usually, the largest part of a text is written in natural language. Any thought can be expressed in natural language. However, sometimes seeking to make the text shorter, clearer or for other reasons we insert certain pieces of data that belong to other semiotic systems, e.g., numbers, dates, abbreviations, which are not natural language. In \cite{Sproat01} these items are denoted as Non-Standard Words (NSW). The main difference between NSW and natural language words is that letter-to-sound rules cannot be directly applied to NSW, the only way to read them out loud is to convert them to natural language first \cite{Taylor09} p. 93.

The present paper deals with text preparation for a Lithuanian speech synthesis system. A set of regular expression-based rules capable of identifying text fragments belonging to other semiotic classes and replacing them with natural language is presented.

The work will be further organized as follows: related works are analyzed in Section~\ref{s2}, theoretical part of regular expressions is given in  Section~\ref{s3}, Section~\ref{s4} introduces the taxonomy of the semiotic classes of the Lithuanian language, Section~\ref{s5} describes the data used in the experiments, Section~\ref{s6} presents regular expressions by semiotic classes written for the expansion of NSWs, Section~\ref{s7} deals with experiments and their results, and Section~\ref{s8} analyses errors.

\section{Related works}\label{s2} 
For the first time a systematic and comprehensive view on the Text Normalization was given in \cite{Sproat01}. Text Normalization can be divided into tokenization, splitting, classification, disambiguation and verbalization. The first step is to divide the text into the word size tokens based on spaces. Composite tokens should be split into sub-tokens if their parts belong to different semiotic classes, e.g., \textit{OpenGL} should be split into \textit{Open GL}. The classification step is to check if the token is a natural language word, otherwise the token is NSW and should be assigned to a certain semiotic class. When talking about NSW, the problem of ambiguity is even greater than in natural language. E.g., it is not clear whether 2018 is a cardinal number or a part of a date, thus disambiguation is necessary. The final step is verbalization, i.e., NSWs are replaced with natural language words. The verbalization method is dependent on the semiotic class NSW belongs to. The taxonomy of semiotic classes was first introduced in \cite{Sproat01}, classification and verbalization rules were also given. Later this taxonomy was revised and extended by other authors \cite{Esch17, Flint17}.

It is important to note that Text Normalization is language dependent, i.e., the taxonomy of semiotic classes, classification and verbalization rules should be written for a particular language. As many as 12 languages are dealt with in \cite{Cho10}: en-GB, de-DE, fr-FR, it-IT, pt-PT, ca-ES, es-ES, nb-NO, da-DK, nl-NL, sv-SE and ko-KR. Additional problems arise in inflectional languages because grammatical forms of expanded words should match the context. E.g., Text Normalization for the Greek language was investigated in \cite{Xydas04}, for Russian in \cite{Cherepanova17}, for Croatian in \cite{Beliga11}, for Macedonian in \cite{Gerazov11}, for Polish in \cite{Brocki12}.

Besides, Text Normalization is task or domain dependent, e.g., normalizing of SMS messages is analysed in \cite{Sikdar22}, whereas Twitter messages are dealt with in \cite{Hanafiah17}.

Various algorithms can be used for text normalization: weighted finite-state transducers (WFSTs) \cite{Ebden14}, Hidden Markov Model (HMM) \cite{Choudhury07}, a phrase-based statistical Machine Translation (MT) model \cite{Aw06}, a character-level MT \cite{Pennell11}, Recurrent Neural Networks (RNN) \cite{Sproat17}. Using regular expressions is not a new idea, normalization is generally done by cascades of simple regular expression substitutions \cite{Jurafsky21}.

It is also noteworthy that the developed text normalization module can be applied not only in text-to-speech synthesis but also to generate large amounts of normalized data, which can then be used to train other normalization models, such as neural networks. As observed in \cite{Zhang19} p. 298, in machine translation large amounts of text and corresponding translated text naturally occur when translators translate texts and distribute them. Meanwhile, there is no reason why people would waste time creating normalized texts, i.e. some effort must be made to create normalized texts for training algorithms.

The normalization of the Lithuanian text has already been examined in \cite{Utka16}, however, their work deals with the morphological analysis of social network comments. Most of the problems are caused by a lack of diacritical marks, and the spellchecking algorithm is used to normalize the text. Identification of semiotic classes was not considered in their work.

Normalization of Lithuanian numbers as a separate task was analyzed in \cite{Balciunas19}. Skip-chain conditional random fields were used for this purpose.

\section{Regular expressions theory}\label{s3} 
In this work, regular expressions (regex for short) were used to detect non-standard words in the text. For more details on these, see, for example, \cite{Jurafsky21} Chapter 2 or \cite{Chodnicki19}. The following is a brief overview of regular expressions, and more specifically, only those elements of regular expressions that were used in the work.

Formally, a regular expression is an algebraic notation for characterizing a set of strings. Regexes are particularly useful in search and replace operations in texts if we have a pattern to search for.

The most basic building block in a regular expression is a character. Most characters in a regex pattern have no special meaning, they simply match themselves e.g., \textit{aBc}. Regular expressions are case sensitive. Some characters have a special meaning. They must be escaped by a backslash if they are meant to represent themselves: $\backslash$\^{}, $\backslash$\$, $\backslash\backslash$, $\backslash$\{, $\backslash$\}, $\backslash$[, $\backslash$], $\backslash$(, $\backslash$), $\backslash$., $\backslash$*, $\backslash$+, $\backslash$?, $\backslash$|.

The string of characters inside square brackets specifies a disjunction of characters to match, e.g., [\textit{aA}]. In cases where there is a well-defined sequence associated with a set of characters, square brackets can be used with the hyphen (-) to specify any character in a range, e.g., [\textit{a-z}], [\textit{A-Z}]. Lithuanian letters do not fall within these ranges, Lithuanian lowercase/uppercase letters are indicated as follows: [\textit{a-ząčęėįšųūž}], [\textit{A-ZĄČĘĖĮŠŲŪŽ}]. In square brackets it is possible to specify to whom the symbol should not be equal, for this purpose a caret (\^{}) is written after the opening square bracket, e.g., [\^{}\textit{a-z}].

There are several predefined character classes:

. - any character except for line endings,
    
$\backslash d$ - a digit: [\textit{0-9}],
    
$\backslash s$ - a whitespace character including tabs and line endings,
    
$\backslash w$ - a word character: [\textit{a-zA-Z\_0-9}].

There are anchor characters that do not match a character as such, they match a boundary:

\^{} - the beginning of a line,
    
\$ - the end of a line,
    
$\backslash b$ - a word boundary.

Quantifiers help specify the expected number of times a pattern will match. The quantifier is written immediately after the character or a group of characters:

+ - one or more occurrences,
    
? - exactly zero or one occurrence,
    
\{n\} - n occurrences,
    
\{n,m\} - from n to m occurrences.

If the regex must match one of several alternative strings, the latter are separated by a pipe symbol, e.g., \(cat|dog\).

Parentheses can be used to group characters. There are several ways to use groups:
\begin{enumerate}
    \item Apply a quantifier to a group, e.g., the regex \("(ab)\textit{+}"\) will match the string \textit{ababab}.
    
    \item Group alternatives, e.g., the regex \("entit(y|ies)"\) will match both \textit{entity} and \textit{entities}.
    
    \item Capture a group, then the group can be used after the \$ character with its number, e.g., the regex \("the\ ([a\textit{-}z]\textit{+})er\ they\ ([a\textit{-}z]\textit{+}),\ the\ \$1er\ we\ \$2"\) will match the string \textit{the faster they ran, the faster we ran}.
    
    \item Use the groups for a replacement, e.g., the regex \("(p\backslash .)\ (\backslash d)"\) will capture two groups that can be used by another regex \("\$2\ \$1"\) which will change the string \textit{p. 7} to \textit{7 p.}
\end{enumerate}

In the present work the rules that contain two regular expressions will be used. If a piece of the string matches the regex on the left side of the rule, it is replaced with the regex on the right side of the rule, e.g.:
\begin{equation}
"(p\backslash .)\ (\backslash d)" \rightarrow "\$2\ \$1". \nonumber
\end{equation}

Finally, a few comments on the limitations of regular expressions. Regular expressions are not recursive. It is possible to construct nested structures to a certain depth, but it is not possible to construct regular expressions for any depth. For example, abbreviations without vowels. They can be identified as a whole but cannot be spelled out. This can only be done if we limit the length of the abbreviations that we can handle.

Another disadvantage of regular expressions is that they are not suitable for validating the values of the identified elements. For example, if we wanted to check whether two years are consecutive, e.g., in the string \textit{School year 2013/2014}, we would get a rather complex structure.

\section{Taxonomy of Lithuanian semiotic classes}\label{s4} 
The taxonomy of Lithuanian non-standard words which is mainly based on the taxonomy presented in \cite{Sproat01}, is shown in Table~\ref{tbl-1}. For a more detailed explanation of semiotic classes see Section~\ref{s6}. The processing of misspelled words (class MISSP) was not considered in this work.

\begin{table}
\caption{Taxonomy of non-standard words in Lithuanian.}\label{tbl-1}
\begin{tabular}{ l l l l }
\hline
& Code & Explanation & Examples \\
\hline
Letters & EXPN & expand	& liet. \(\rightarrow\) lietuviškai \\
& LSEQ & letter sequence & VU \(\rightarrow\) vė-u \\ 
& ASWD & read as word & KAM \\
\hline
Numbers & NUM & cardinal number & 10 km. \\
& NORD & ordinal number & XIX a., 1941-ųjų \\
& NTEL & telephone & 8-611-99999 \\
& NTIME & time & 10:45 \\
& NDATE & date & 2018 m. spalio 15 d., 2018 10 15 \\
& NYEAR & year & 1941 m. \\
& NCODE & code & ISBN: 1909232424 \\
\hline
Other & MISSP & misspelled word &  liett. \\
& URL & url, pathname or email & pkasparaitis@yahoo.com \\
& NONE & should be ignored & ascii art, \&nbsp; \\
\hline
\end{tabular}
\end{table}

\section{Data used}\label{s5} 
Three widely used and as diverse as possible datasets were chosen: 1) data from the news portal Delfi (\url{https://www.delfi.lt/}). Delfi is a major internet portal in Lithuania providing daily news from politics to culture and sports. It is one of the most popular websites among Lithuanian users. The data were collected in October 2018. 2) Data from the website of the Faculty of Philology of Vilnius University (\url{https://www.flf.vu.lt/}) collected during the same period. This is a typical website of a university, school, or similar institution. 3) Data generated by iGO Primo Nextgen car navigation. The navigation is fully Lithuanized, its skin (Skin iGO NextGen by pongo) currently is available at \url{https://www.gpspower.net/igo-primo-nextgen-skins.html} (last viewed October 20, 2022, registration required). The data were collected during several trips from Lithuania to Poland, Germany and France. These three datasets are further abbreviated as DEL, FLF and NAV. The number of words in the datasets, as well as the number of non-standard words by semiotic classes, are given in Table~\ref{tbl-2}.

\begin{table}
\caption{Distribution of non-standard words in datasets.}\label{tbl-2}
\begin{tabular}{ l l l l l }
\hline
& Code & DEL & FLF & NAV \\
\hline
Letters & EXPN & 812 (15.4\%) & 1760 (16.4\%) & 3785 (35.6\%) \\
& LSEQ & 1029 (19.5\%) & 2859 (26.6\%) & 448 (4.2\%) \\
& ASWD & 1246 (23.6\%) & 2048 (19.0\%) & 1599 (15.0\%) \\
\hline
Numbers & NUM & 1440 (27.2\%) & 766 (7.1\%) & 4545 (42.7\%) \\
& NORD & 243 (4.6\%) & 658 (6.1\%) & 35 (0.3\%) \\
& NTEL & - & 66 (0.6\%) & - \\
& NTIME & 35 (0.7\%) & 502 (4.7\%) & 5 (0\%) \\
& NDATE & 102 (1.9\%) & 850 (7.9\%) & - \\
& NYEAR & 339 (6.4\%) & 948 (8.8\%) & - \\
& NCODE & - & 127 (1.2\%) & 49 (0.5\%) \\
\hline
Other & MISSP & 3 (0\%) & 30 (0.3\%) & - \\
& URL & 38 (0.7\%) & 142 (1.3\%) & - \\
& NONE & 1 (0\%) & 8 (0.1\%) & 173 (1.6\%) \\
\hline
\multicolumn{2}{ l }{Total non-standard words} & 5288 (5.3\%) & 10764 (10.6\%) & 10639 (21.5\%) \\
\multicolumn{2}{ l }{Standard words} & 94838 & 90540 & 38884 \\
\multicolumn{2}{ l }{Total words} & 100126 & 101298 & 49523 \\
\hline
\end{tabular}
\end{table}

Table~\ref{tbl-2} shows that the first and second datasets contain over 100,000 words, and the third one contains nearly 50,000 words. The proportion of NSW in the datasets varies greatly, with over 5\% in the first (DEL) dataset, over 10\% in the second (FLF) and over 20\% in the third (NAV). In all sets, a significant part of NSW consists of EXPN, LSEQ, ASWD, and NUM. The FLF set also contains a number of NSW related to the date and time. The DEL set contains neither NTEL nor NCODE, and the NAV set does not contain NSWs of the following classes: NTEL, NDATE, NYEAR, URL. The absolute majority in the NAV set consists of NUM and EXPN class NSWs, as it abounds in examples of such a type: \textit{Už 1 km 700 m (After 1 km 700 m), Iki kitos nuorodos 4 km 800 m. (Up to another reference 4 km 800 m), Greičio ribojimas 50 km/h (Speed limit 50 km/h)}.

NSWs were manually selected from three datasets along with a particular context and listed in three data tables. The context was taken so as to help determine the correct grammatical form of the NSW being expanded. If there are several interconnected NSWs, they are taken as one entry up to 4 NSWs per entry, but the vast majority of entries contain only one NSW. NSW classes are listed next to each entry. The correct expansion of the entry is also indicated there. An example of a data table is shown in Table~\ref{tbl-3}.

\begin{table}
\caption{Example of a data table.}\label{tbl-3}
\begin{tabular}{ l l l l l l l l }
\hline
No. & Entry & NSW1 & NSW2 & NSW3 & NSW4 & Correct expansion & Generated expansion \\
\hline
1. & Nuo 2000 m. & NYEAR & EXPN & - & - & Nuo dutūkstantųjų metų & \\
2. & M1 & LSEQ & NUM & - & - & Em vienas & \\
3. & ... & & & & & & \\
\hline
\end{tabular}
\end{table}

\section{Rules implemented}\label{s6} 
A set of rules has been created. When normalizing a text, the rules are applied sequentially, i.e., the first rule is applied to all substrings it matches, followed by the second rule applied, and so on. The rules are put in the following order so that: 1) rules matching a longer substring are executed first, e.g., dates before years. 2) some rules can produce data for the next rules, e.g., dates or long numbers can be split into smaller numbers. Another idea used was as follows: some rules insert special tags before or after NSW. These tags denote features such as gender, singular/plural, ordinal/cardinal, etc. These tags are used by the following rules to choose the correct word form. Besides, some rules can duplicate these tags if the NSW is split into multiple NSWs.

Below we will discuss the rules according to semiotic classes.

\subsection{Cardinal numerals and units of measure}
Let us start with cardinal numerals because expanding them is one of the most difficult tasks, and they are also used as an integral part of other more complex expressions.

The numeral is often preceded by a preposition, which describes the inflection of the numeral. Without going deep into which prepositions of the Lithuanian language are used with numbers and which are not, let us take all the prepositions of the Lithuanian language. The following 44 prepositions require the genitive case: \textit{anot, ant, arti, aukščiau, be, dėka, dėl, dėlei, greta, iki, lig, ligi, iš, link, linkui, netoli, nuo, pasak, pirmiau, pirm, prie, pusiau, šalia, tarp, toliau, žemiau, vidury, vidur, vietoj, virš, viršum, viršuj, išilgai, įstrižai, įkypai, skersai, kiaurai, skradžiai, abipus, anapus, šiapus, abigaliai, iš po, iš už}. E.g., \textit{Nuo 21 min. iki 2 val. \(\rightarrow\) Nuo dvidešimt vienos minutės iki dviejų valandų (From twenty-one minute to two hours.)}. The following 13 prepositions take the accusative case: \textit{apie, aplink, aplinkui, į, pagal, palei, pas, paskui, paskum, per, prieš, priešais, pro}. E.g., \textit{Per 21 min. \(\rightarrow\) Per dvidešimt vieną minutę (In twenty-one minutes.)}. The prepositions \textit{su, sulig, ties} take the instrumental case. The preposition \textit{už} requires the genitive case when talking about time and distance, e.g., \textit{Už 2 km. \(\rightarrow\) Už dviejų kilometrų (Two kilometers away.)}. This can be determined by the unit of measure. Let us assume that in other cases the preposition \textit{už} takes the accusative case, e.g., \textit{Už 2 Lt. \(\rightarrow\) Už du litus (For two litas.)}. Similarly the preposition \textit{po} requires the genitive case when talking about time, e.g., \textit{Po 2 val. \(\rightarrow\) Po dviejų valandų (In two hours.)}. Let us assume that in other cases the preposition \textit{po} requires the accusative case, e.g., \textit{Po 2 Lt. \(\rightarrow\) Po du litus (Two litas each.)} despite the fact that the instrumental can also be used with this preposition, e.g., \textit{Po septyniais užraktais (Under seven locks.)}, however, in these cases the numbers are usually written in plain text \cite{GML96} p. 437-455.

The rules that insert a special tag indicating the case of the numeral were written. The following symbols will be used in special tags: \textit{S} – singular, \textit{P} – plural, \textit{N} – nominative, \textit{G} – genitive, \textit{A} – accusative, \textit{I} – instrumental, \textit{F} – feminine, \textit{M} – masculine, \textit{O} – ordinal, \textit{C} – cardinal. Tags are enclosed between \textit{«} and \textit{»} and can contain multiple symbols, e.g., \textit{«PG»}, \textit{«FO»}. We will use plural where singular/plural distinction makes no difference. For example, the rule 
\begin{equation}
"\backslash b([Ss]u|[Ss]ulig|[Tt]ies)\ (\backslash d)" \rightarrow "\$1\ \textit{«PI»}\ \$2" \nonumber
\end{equation}

will expand the string \textit{su 5} into \textit{su «PI» 5}. Sometimes a single preposition requires the whole sequence of numerals and units of measure arranged in descending order to be used in a certain case, e.g., \textit{Už 2 val. 15 min. 30 sek. Už 3 km 500 m. (In 2 h 15 min 30 s. 3 km 500 m away.)}. Such structures are badly needed in navigation. Rules that duplicate a special tag before each numeral were written. E.g., the rule
\begin{equation}
"(\textit{«}[SP][GAI]\textit{»})\ (\backslash d\textit{+})\ (km)\ (\backslash d\textit{+})\ (m)\backslash b" \rightarrow "\$1\ \$2\ \$3\ \$1\ \$4\ \$5" \nonumber
\end{equation}

will expand the string \textit{«PG» 5 km 300 m} into \textit{«PG» 5 km «PG» 300 m}. Now, each pair of numerals and units of measure can be further processed separately.
It is time to split large numbers into 3-digit groups. The following rules were used for this purpose:
\begin{equation}\label{eq-1}
"(\textit{«}[SP][GAI]\textit{»})\ (\backslash d\{1,12\})\backslash b" \rightarrow "\$1\ \$2\ \$1",
\end{equation}
\begin{equation}\label{eq-2}
\begin{array}{l}
"(\textit{«}[SP][GAI]\textit{»})\ (\backslash d\{1,3\})(\backslash d\{9\})\backslash b" \rightarrow "\$1\ \$2\ \$1\ mlrd\ \$1\ \$3", \\
"(\textit{«}[SP][GAI]\textit{»})\ (\backslash d\{1,3\})(\backslash d\{6\})\backslash b" \rightarrow "\$1\ \$2\ \$1\ mln\ \$1\ \$3", \\
"(\textit{«}[SP][GAI]\textit{»})\ (\backslash d\{1,3\})(\backslash d\{3\})\backslash b" \rightarrow "\$1\ \$2\ \$1\ t\textit{ū}kst\ \$1\ \$3".
\end{array}
\end{equation}

The rule (\ref{eq-1}) inserts a special tag before the unit of measure. After the rules (\ref{eq-1}) and (\ref{eq-2}) have been successively applied to the string \textit{«PA» 123456789234 km}, the following result will be obtained: \textit{«PA» 123 «PA» mlrd «PA» 456 «PA» mln «PA» 789 «PA» tūkst «PA» 234 «PA» km}. It should be noted that abbreviations of billion, million and thousand are used because the exact grammatical form is unknown at this stage. Besides, this notation allows such abbreviations already present in the text to be expanded, e.g., \textit{Su 100 mln. \(\rightarrow\) Su šimtu milijonų}. For simplicity, they will be treated as units of measure further in this work.
In order not to overload the paper with small details, we are not going to explain the 3-digit expansion in detail; we will present only a few most important facts:
\begin{itemize}
    \item If the number ends with 1 (not with 11), the unit of measure should be written in singular, otherwise in plural, for example, \textit{21 metras (singular), 22 metrai (plural)};
    
    \item If the unit of measure is feminine, the last digit is also feminine, otherwise it is masculine, for example, \textit{2 s \(\rightarrow\) dvi sekundės (feminine), 2 m \(\rightarrow\) du metrai (masculine)};
    
    \item Numbers ending with 11, 12, ..., 19, as well as numbers ending with 0 (for example, 10, 20, 90, 100), require a unit of measure to be written in the genitive case, and this requirement overrides the case defined by the preposition, however, it does not move to the next number groups. For example, \textit{Prieš 113003 m.} you need to expand as follows: \textit{Prieš šimtą (accusative) trylika (acc.) tūkstančių (gen.) tris (acc.) metus (acc.)}, here the case is indicated in the brackets after the word.
\end{itemize}

Simple rules are used to finish the expansion of numbers, they select appropriate grammatical form of the number based on special tags. The case and singular/plural tags come from the left side while the gender tag comes from the right side. After that, similar rules are used to expand abbreviations \textit{mldr, mln, tūkst} and the units of measure. E.g.:
\begin{equation}
\begin{array}{l}
"\textit{«}PG\textit{»}\ 4\ \textit{«}FC\textit{»}" \rightarrow "keturi\textit{ų}", \\
"\textit{«}PA\textit{»}\ 4\ \textit{«}FC\textit{»}" \rightarrow "keturias", \\
"\textit{«}PI\textit{»}\ 4\ \textit{«}FC\textit{»}" \rightarrow "keturiomis",
\end{array} \nonumber
\end{equation}
\begin{equation}
\begin{array}{l}
"\textit{«}PG\textit{»}\ mln\backslash b" \rightarrow "milijon\textit{ų}", \\
"\textit{«}PA\textit{»}\ mln\backslash b" \rightarrow "milijonus", \\
"\textit{«}PI\textit{»}\ mln\backslash b" \rightarrow "milijonais",
\end{array} \nonumber
\end{equation}
\begin{equation}
\begin{array}{l}
"\textit{«}PG\textit{»}\ km\backslash b" \rightarrow "kilometr\textit{ų}", \\
"\textit{«}PA\textit{»}\ km\backslash b" \rightarrow "kilometrus", \\
"\textit{«}PI\textit{»}\ km\backslash b" \rightarrow "kilometrais".
\end{array} \nonumber
\end{equation}

If there is no preposition before the cardinal numeral, the latter is expanded into the nominative. The number is often followed by an abbreviation of the unit of measure. The gender of the number depends on the gender of the unit of measure. E.g., the feminine gender should be used to expand \textit{3 min. 43,91 sek. \(\rightarrow\) trys minutės keturiasdešimt trys kablelis devyniasdešimt viena sekundės}. If there is no unit of measure, masculine should be used.

\subsection{Ordinal numerals}
Ordinal numerals can be written in several ways: 1) in Arabic numerals, adding an ending through a hyphen, e.g., \textit{2-as ($2^{nd}$)}, 2) in Arabic numerals, e.g., \textit{1 pavyzdys (Example 1)}, 3) in Roman numerals, e.g., \textit{XIX skyrius (Chapter XIX)}.

Ordinal numerals are easy to identify if they are written in \textbf{Arabic numerals, with a hyphen and an ending} which describes their grammatical form. The ending also indicates whether the ordinal number is a pronominal one. Ordinal numbers (including the composite ordinal numbers) are made up of the nominative case of cardinal numbers, only the last word has the form of the ordinal number \cite{GML96} p. 243, e.g., cardinal number \textit{643 \(\rightarrow\) Šeši šimtai keturiasdešimt trys}, ordinal number \textit{643-ias \(\rightarrow\) Šeši šimtai keturiasdešimt trečias}. It should be noted that the endings given in this way have higher priority than the preposition of the number, e.g., the preposition \textit{apie} takes the accusative case, thus \textit{Apie 2004. \(\rightarrow\) Apie du (acc.) tūkstančius (acc.) keturis (acc.)}. However, by indicating the ending of the ordinal number, we get the following: \textit{Apie 2004-ųjų pabaigą. \(\rightarrow\) Apie du (nom.) tūkstančiai (nom.) ketvirtųjų (gen.) pabaigą (acc.)}. So, first, we replace the case defined with the preposition into the nominative case using the following rule:
\begin{equation}
"\textit{«}S([GAI])\textit{»}\ (\backslash d\{1,12\})(\textit{-}[a\textit{-ząčęėįšųūž}]\textit{+}\backslash b)" \rightarrow "\textit{«}SN\textit{»}\ \$2\ \$3". \nonumber
\end{equation}

Then the proper grammatical form of the number is chosen based on the ending specified, e.g., \textit{2004-ųjų \(\rightarrow\) du tūkstančiai 4-ųjų \(\rightarrow\) du tūkstančiai ketvirtųjų}. The following rule is used for this purpose:
\begin{equation}\label{eq-mmm3}
"\backslash bketuri\textit{-ųjų}\backslash b" \rightarrow "ketvirt\textit{ųjų}".
\end{equation}

More sophisticated rules can also be used for this purpose, such as:
\begin{equation}
\begin{array}{l}
"(dvi|tris|keturias|penkias|\textit{šeš}ias|septynias|a\textit{š}tuonias|devynias)de\textit{š}imt\textit{-}oji" \\
\rightarrow "\$1de\textit{š}imtoji".
\end{array} \nonumber
\end{equation}

Sometimes ordinal numerals are written in \textbf{Arabic} numerals, and the word after them suggests that it is an ordinal numeral, such as auditoriums, classrooms, articles are numbered rather than counted, because we usually do not care what their total number is. Roman numerals may not be appropriate for this, as there may be hundreds of classrooms or auditoriums. E.g., \textit{104 aud. \(\rightarrow\) šimtas ketvirta aud., 104A kabinetas \(\rightarrow\) šimtas ketvirtas A kabinetas, 90 str. \(\rightarrow\) devyniasdešimtas str}. In this case, the following rules are used first to insert special tags indicating that it is an ordinal numeral and its gender:
\begin{equation}
\begin{array}{l}
"\backslash b(\backslash d{1,3})([ABC]?\ aud)" \rightarrow "\$1\ \textit{«}FO\textit{»}\ \$2", \\
"\backslash b(\backslash d{1,3})([ABC]?\ kab)" \rightarrow "\$1\ \textit{«}MO\textit{»}\ \$2", \\
"\backslash b(\backslash d{1,3})\ (str)" \rightarrow "\$1\ \textit{«}MO\textit{»}\ \$2".
\end{array} \nonumber
\end{equation}

Expansion is completed by means of the following rules:
\begin{equation}\label{eq-nnn4}
\begin{array}{l}
"\textit{«}SG\textit{»}\ 4\ \textit{«}FO\textit{»}" \rightarrow "ketvirtos", \\
"\textit{«}SA\textit{»}\ 4\ \textit{«}FO\textit{»}" \rightarrow "ketvirt\textit{ą}", \\
"\textit{«}SI\textit{»}\ 4\ \textit{«}FO\textit{»}" \rightarrow "ketvirta", \\
"4\ \textit{«}FO\textit{»}" \rightarrow "ketvirta".
\end{array}
\end{equation}

Attempts were made to adapt \textbf{Roman} numerals to the inflection of the word following it, e.g., \textit{I etapas \(\rightarrow\) pirmas etapas, I kvietimo \(\rightarrow\) pirmo kvietimo, I vieta \(\rightarrow\) pirma vieta, I pakopos \(\rightarrow\) pirmos pakopos, I dalis \(\rightarrow\) pirma dalis, I rūmai \(\rightarrow\) pirmi rūmai, I skyriuje \(\rightarrow\) pirmame skyriuje, I amžiaus \(\rightarrow\) pirmo amžiaus, I amžius \(\rightarrow\) pirmas amžius, I mokslinės konferencijos \(\rightarrow\) pirmos mokslinės konferencijos, I mokslinė konferencija \(\rightarrow\) pirma mokslinė konferencija, I vietą \(\rightarrow\) pirmą vietą}. The following rules are written for this purpose:
\begin{equation}
\begin{array}{l}
"\backslash bI\ ([a\textit{-ząčęėįšųūž}]\textit{+})(as|o|a|os)\backslash b" \rightarrow "pirm\$2\ \$1\$2", \\
"\backslash bI\ ([a\textit{-ząčęėįšųūž}]\textit{+})(is)\backslash b" \rightarrow "pirma\ \$1\$2", \\
"\backslash bI\ ([a\textit{-ząčęėįšųūž}]\textit{+})(ai)\backslash b" \rightarrow "pirmi\ \$1\$2", \\
"\backslash bI\ ([a\textit{-ząčęėįšųūž}]\textit{+})(e)\backslash b" \rightarrow "pirmame\ \$1\$2", \\
"\backslash bI\ ([a\textit{-ząčęėįšųūž}]\textit{+})(aus)\backslash b" \rightarrow "pirmo\ \$1\$2", \\
"\backslash bI\ ([a\textit{-ząčęėįšųūž}]\textit{+})(us)\backslash b" \rightarrow "pirmas\ \$1\$2", \\
"\backslash bI\ ([a\textit{-ząčęėįšųūž}]\textit{+})(\textit{ė}s)\backslash b" \rightarrow "pirmos\ \$1\$2", \\
"\backslash bI\ ([a\textit{-ząčęėįšųūž}]\textit{+})\backslash b(\textit{ė})" \rightarrow "pirma\ \$1\$2", \\
"\backslash bI\ ([a\textit{-ząčęėįšųūž}]\textit{+})\backslash b(\textit{ą})" \rightarrow "pirm\$2\ \$1\$2".
\end{array} \nonumber
\end{equation}

Analogous rules are written for other Roman numerals up to 30.
In general, Roman numerals are expanded into the nominative case singular. E.g., \textit{III. Antikos istorijos \(\rightarrow\) trečias. Antikos istorijos}. It is important to make sure that it is not a part of a larger element (e.g., \textit{III} is not a part of \textit{XIII}). For Roman numerals \textit{I} and \textit{V}, it remains to be checked if this is not the first letter of the word capitalized. In addition, they can be used with a period, in this case they can be confused with initials. It was decided to always treat \textit{I.} and \textit{V.} as initials, as only in the FLF data \textit{I.} had to be expanded twice (out of 16) as a Roman numeral, and in all other cases \textit{I.} and \textit{V.} meant initials.
\begin{equation}
\begin{array}{l}
"(\hat{\ }|[\backslash s\backslash (])I([\hat{\ }\backslash w\backslash .\textit{ąčęėįšųūž}]|\$)" \rightarrow "\$1pirmas\$2", \\
"(\hat{\ } |[\backslash s\backslash (])II([\hat{\ }\backslash w]|\$)" \rightarrow "\$1antras\$2", \\
"(\hat{\ }|[\backslash s\backslash (])III([\hat{\ }\backslash w]|\$)" \rightarrow "\$1tre\textit{č}ias\$2", \\
"(\hat{\ }|[\backslash s\backslash (])IV([\hat{\ }\backslash w]|\$)" \rightarrow "\$1ketvirtas\$2", \\
"(\hat{\ }|[\backslash s\backslash (])V([\hat{\ }\backslash w\backslash .\textit{ąčęėįšųūž}]|\$)" \rightarrow "\$1penktas\$2", \\
"(\hat{\ }|[\backslash s\backslash (])VI([\hat{\ }\backslash w]|\$)" \rightarrow "\$1\textit{šeš}tas\$2".
\end{array} \nonumber
\end{equation}

There were cases where the ending was also indicated for a Roman numerals, e.g., \textit{I-ieji rūmai \(\rightarrow\) pirmas-ieji rūmai \(\rightarrow\) pirmieji rūmai}. In that case, special rules were also created to correct the grammatical form:
\begin{equation}
"\backslash bpirmas\textit{-}ieji\backslash b" \rightarrow "pirmieji". \nonumber
\end{equation}

Note that an ordinal rather than a cardinal numeral is adjusted here, unlike the rule (\ref{eq-mmm3}).

Roman numerals with the abbreviation \textit{a.} were processed in a special way. In data set FLF, Roman numerals \textit{I-IV a.} were expanded into \textit{aukštas (floor)}, e.g., \textit{III a. \(\rightarrow\) trečias aukštas}, larger ones – into \textit{amžius (century)}. In dataset DEL, the abbreviation \textit{a.} was always expanded into \textit{amžius}. The abbreviation \textit{a.} before the lowercase letters was expanded into the word \textit{amžius} in the genitive singular, e.g., \textit{XIX a. pradžioje \(\rightarrow\) devyniolikto amžiaus pradžioje}, and at the end of the sentence into one in the nominative singular, e.g., \textit{Lietuvių literatūros istorija, XIX a. \(\rightarrow\) Lietuvių literatūros istorija, devynioliktas amžius}.
\begin{equation}
\begin{array}{l}
"\backslash bIII\ a\backslash ." \rightarrow "tre\textit{č}ias\ auk\textit{š}tas", \\
"\backslash b(XIX\ a\backslash .)\ ([a\textit{-}z])" \rightarrow "devyniolikto\ am\textit{ž}iaus\ \$2", \\
"\backslash bXIX\ a\backslash ." \rightarrow "devynioliktas\ am\textit{ž}ius".
\end{array} \nonumber
\end{equation}

Special rules were created for the following constructions: \textit{III/IV d. \(\rightarrow\) trečia dalis iš keturių (third part of four), III/IV \(\rightarrow\) trečia iš keturių}.
\begin{equation}
\begin{array}{l}
"\backslash bIII/IV\ d\backslash ." \rightarrow "tre\textit{č}ia\ dalis\ i\textit{š}\ keturi\textit{ų}", \\
"\backslash bIII/IV\backslash b" \rightarrow "tre\textit{č}ia\ i\textit{š}\ keturi\textit{ų}".
\end{array} \nonumber
\end{equation}

\subsection{Dates}
There are two date formats in Lithuanian \cite{Vladarskiene22}, p. 203: long and short. In the long format, the year and day are written in numbers, and the month is written in words, e.g., \textit{2013 m. sausio 4 d.} In the short format, the entire date is written in Arabic numerals, the year is four-digit, the month and day are two-digit separated by spaces or hyphens. If the month or day is one-digit, it is preceded by zero, e.g., \textit{2013 01 04}. 

First, some words about expanding the \textbf{long format date}. The year and day are replaced with ordinal numbers. The date can be preceded by a preposition. The year and month should always be said in the genitive case, the preposition does not affect their case, it defines the case of the day. The day is used in the singular.

First of all, it is necessary to identify the date, and then it can be changed to a template, the processing of the individual parts thereof being already known (to us). Instead of a full list of the months \textit{sausio| vasario| kovo| balandžio| gegužės| birželio| liepos| rugpjūčio| rugsėjo| spalio| lapkričio| gruodžio}, for the sake of brevity, we will only write the first and the last \textit{sausio|...|gruodžio}. The following rule was used for long format dates:
\begin{equation}
\begin{array}{l}
"\textit{«}P([GAI])\textit{»}\ ([12]\backslash d\backslash d\backslash d)\ m\backslash .\ (sausio| ... |gruod\textit{ž}io)\ (\backslash d?\backslash d)\ (d\backslash .|dien)" \\
\rightarrow "\$2\textit{-ų}\ met\textit{ų}\ \$3\ \textit{«}S\$1\textit{»}\ \$4\ \textit{«}FO\textit{»}\ \textit{«}S\$1\textit{»}\ \$5\ \textit{«}S\$1\textit{»}".
\end{array} \nonumber
\end{equation}

We finalize the expansion of the date using the above-mentioned rules for expansion of ordinal numbers, group of rules (\ref{eq-nnn4}), as well as the following rules to expand the abbreviation \textit{d.}:
\begin{equation}
\begin{array}{l}
"\textit{«}SG\textit{»}\ d\backslash ." \rightarrow "dienos", \\
"\textit{«}SA\textit{»}\ d\backslash ." \rightarrow "dien\textit{ą}", \\
"\textit{«}SI\textit{»}\ d\backslash ." \rightarrow "diena".
\end{array} \nonumber
\end{equation}

E.g., \textit{Nuo 2013 m. sausio 4 d. \(\rightarrow\) Nuo «PG» 2013 m. sausio 4 d. \(\rightarrow\) Nuo 2013-ų metų sausio «SG» 4 «FO» «SG» d. «SG» \(\rightarrow\) Nuo du tūkstančiai tryliktų metų sausio ketvirtos dienos «SG»}.

The tag of the inflection specified by the preposition is passed on beyond the date where it can be used to define the inflection of time, e.g., \textit{Nuo 2013 m. sausio 4 d. 15 val}. If the word \textit{diena (day)} is used instead of the abbreviation \textit{d.}, the word \textit{diena} is left unchanged.

If there is no preposition before the date, the day is expanded into the accusative case singular, e.g., \textit{2013 m. sausio 14 d. įvyko \(\rightarrow\) du tūkstančiai tryliktų metų sausio keturioliktą dieną įvyko}.
\begin{equation}
\begin{array}{l}
"\backslash b([12]\backslash d\backslash d\backslash d)\ m\backslash .\ (sausio| ... |gruod\textit{ž}io)\ (\backslash d?\backslash d)\ (d\backslash .|dien)" \\
\rightarrow "\$1\textit{-ų}\ met\textit{ų}\ \$2\ \textit{«}SG\textit{»}\ \$3\ \textit{«}FO\textit{»}\  \textit{«}SG\textit{»}\ \$4".
\end{array} \nonumber
\end{equation}

There were 5 samples with the abbreviation \textit{m.} omitted, e.g., \textit{2018 sausio 5 d. baigiasi terminas}. Although this is an irregular format, a separate rule without an abbreviation \textit{m.} was created for them.
\begin{equation}
\begin{array}{l}
"\textit{«}P([GAI])\textit{»} ([12]\backslash d\backslash d\backslash d)\ (sausio| ... |gruod\textit{ž}io)\ (\backslash d?\backslash d)\ (d\backslash .|dien)" \\
\rightarrow "\$2\textit{-ų}\ \$3\ \textit{«}S\$1\textit{»}\ \$4\ \textit{«}FO\textit{»}\ \textit{«}S\$1\textit{»}\ \$5\ \textit{«}S\$1\textit{»}".
\end{array} \nonumber
\end{equation}

The range of days can be specified in long format dates, e.g., \textit{2013 m. sausio 14–15 d. \(\rightarrow\) du tūkstančiai tryliktų metų sausio keturioliktą penkioliktą dienomis}. Days should be separated by a dash without spaces, however, there were cases where the days were separated by a hyphen and spaces were left. Days are expanded into the accusative case singular, the abbreviation \textit{d.} is expanded into the instrumental case plural. If there is a preposition, it is ignored. A separate rule was created for the case where the abbreviation \textit{m.} was missing.
\begin{equation}
\begin{array}{l}
"\backslash b([12]\backslash d\backslash d\backslash d)\ m\backslash .\ (sausio| ... |gruod\textit{ž}io)\ (\backslash d?\backslash d)\ ?[-\ \textit{-}]\ ?(\backslash d?\backslash d)\ ?(d\backslash .|dien)" \\
\rightarrow "\$1\textit{-ų}\ met\textit{ų}\ \$2\ \textit{«}SA\textit{»}\ \$3\ \textit{«}FO\textit{»}\ \textit{«}SA\textit{»}\ \$4\ \textit{«}FO\textit{»}\  \textit{«}PI\textit{»}\ \$5".
\end{array} \nonumber
\end{equation}

There were long format dates with no year specified. If there was a preposition before that date, it was necessary to use an additional rule to insert a special tag, as that date did not begin with a number, e.g., \textit{Nuo sausio 14 d. \(\rightarrow\) Nuo «PG» sausio 14 d.} 
\begin{equation}\label{eq-mmm1}
\begin{array}{l}
"\backslash b([Aa]rti|[Ii]ki|[Ll]igi|[Nn]uo|[Tt]arp|[Vv]ietoj|[Uu]\textit{ž}|[Pp]o)\  \\
(sausio| ... |gruod\textit{ž}io)\backslash b" \rightarrow 
"\$1\ \textit{«}PG\textit{»}\ \$2".
\end{array}
\end{equation}

The expansion of the date itself is performed using the second rule:
\begin{equation}\label{eq-mmm2}
\begin{array}{l}
"\textit{«}P([GAI])\textit{»}\ (sausio| ... |gruod\textit{ž}io)\ (\backslash d?\backslash d)\ (d\backslash .|dien)" \\
\rightarrow "\$2\ \textit{«}S\$1\textit{»}\ \$3\ \textit{«}FO\textit{»}\  \textit{«}S\$1\textit{»}\ \$4\ \textit{«}S\$1\textit{»}".
\end{array}
\end{equation}

There may also be long format dates that do not specify a year and there is no preposition. The sentence can start with such a date, so the month can be capitalized. The accusative case singular is used, e.g., \textit{Sausio 14 d. \(\rightarrow\) Sausio keturioliktą dieną}.
\begin{equation}
\begin{array}{l}
"\backslash b([sS]ausio| ... |[gG]ruod\textit{ž}io)\ (\backslash d?\backslash d)\ (d\backslash .|dien)" \\
\rightarrow "\$1\ \textit{«}SA\textit{»}\ \$2\ \textit{«}FO\textit{»}\ \textit{«}SA\textit{»}\ \$3".
\end{array} \nonumber
\end{equation}

There can be long format dates without years, with a range of days, e.g., \textit{sausio 14–15 d. \(\rightarrow\) sausio keturioliktą penkioliktą dienomis}. Days are expanded into the accusative case singular, the abbreviation \textit{d.} is expanded into the instrumental case plural. If there is a preposition, it is ignored.
\begin{equation}
\begin{array}{l}
"\backslash b([sS]ausio| ... |[gG]ruod\textit{ž}io)\ (\backslash d?\backslash d)\ ?[-\ \textit{-}]\ ?(\backslash d?\backslash d)\ ?(d\backslash .|dien)" \\
\rightarrow "\$1\ \textit{«}SA\textit{»}\ \$2\ \textit{«}FO\textit{»}\  \textit{«}SA\textit{»}\ \$3\ \textit{«}FO\textit{»}\ \textit{«}PI\textit{»}\ \$4". 
\end{array} \nonumber
\end{equation}

The long format date may not specify a day, but may have the month abbreviation mėn. If there is a preposition, it defines the grammatical form of the abbreviation \textit{mėn.}, e.g., \textit{iki 2013 m. sausio mėn. \(\rightarrow\) iki du tūkstančiai tryliktų metų sausio mėnesio}.
\begin{equation}
\begin{array}{l}
"\textit{«}P([GAI])\textit{»}\ ([12]\backslash d\backslash d\backslash d)\ m\backslash .\ (sausio| ... |gruod\textit{ž}io)\ m\textit{ė}n\backslash ." \\
\rightarrow "\$2\textit{-ų}\ met\textit{ų}\ \$3\ \textit{«}S\$1\textit{»}\ m\textit{ė}n.".
\end{array} \nonumber
\end{equation}

If there is no preposition, the abbreviation \textit{mėn.} is expanded into the accusative case singular, e.g., \textit{2013 m. sausio mėn. \(\rightarrow\) du tūkstančiai tryliktų metų sausio mėnesį}.
\begin{equation}
\begin{array}{l}
"\backslash b([12]\backslash d\backslash d\backslash d)\ m\backslash .\ (sausio| ... |gruod\textit{ž}io)\ m\textit{ė}n\backslash ." \\
\rightarrow "\$1\textit{-ų}\ met\textit{ų}\ \$2\ \textit{«}SA\textit{»}\ m\textit{ė}n."
\end{array} \nonumber
\end{equation}

Later the following rules are used to expand the abbreviation \textit{mėn.}:
\begin{equation}
\begin{array}{l}
"\textit{«}SG\textit{»}\ m\textit{ė}n\backslash ." \rightarrow "m\textit{ė}nesio", \\
"\textit{«}SA\textit{»}\ m\textit{ė}n\backslash ." \rightarrow "m\textit{ė}nes\textit{į}", \\
"\textit{«}SI\textit{»}\ m\textit{ė}n\backslash ." \rightarrow "m\textit{ė}nesiu".
\end{array} \nonumber
\end{equation}

If the abbreviation \textit{mėn.} is absent, the year is usually expanded into the genitive case plural, the month is usually specified in the nominative case singular, e.g., \textit{2013 m. sausis \(\rightarrow\) du tūkstančiai tryliktų metų sausis}.
\begin{equation}
"\backslash b([12]\backslash d\backslash d\backslash d)\ m\backslash .\ (sausis| ... |gruodis)" \rightarrow "\$1\textit{-ų}\ met\textit{ų}\ \$2". \nonumber
\end{equation}

Another common construction is similar to a long format date, but uses the abbreviation \textit{š. m.} or full text \textit{šių metų (this year)} instead of the year, which may be preceded by a preposition, e.g., \textit{Iki š. m. sausio 14 d., iki šių metų sausio 14 d.} In this case, it is first necessary to insert a special tag after the preposition, just as in (\ref{eq-mmm1}).
\begin{equation}
"([Nn]uo|[Ii]ki|[Ll]igi)\ (\textit{š}\backslash .\ ?m\backslash .|\textit{šių}\ met\textit{ų})" \rightarrow "\$1\ \$2\ \textit{«}PG\textit{»}". \nonumber
\end{equation}

Then the rule (\ref{eq-mmm2}) can be applied. The abbreviation \textit{š. m.} is expanded using the following rule:
\begin{equation}
"\backslash b([\textit{šŠ}])\backslash .\ ?m\backslash ." \rightarrow "\$1i\textit{ų}\ met\textit{ų}". \nonumber
\end{equation}

In order to expand the date written in a \textbf{short form}, we use the following rule first:
\begin{equation}
\begin{array}{l}
"\textit{«}P([GAI])\ ([12]\backslash d\backslash d\backslash d)[\ \textit{-}](0[1\textit{-}9]|10|11|12)[\ \textit{-}]([0\textit{-}3]\backslash d)\backslash b" \\
\rightarrow "\$2\textit{-ų}\ met\textit{ų}\ \textit{«}\$3m\textit{ė}n\textit{»}\ \textit{«}S\$1\textit{»}\ \$4\ \textit{«}FO\textit{»}\ \textit{«}S\$1\textit{»}\ d."
\end{array} \nonumber
\end{equation}

We get the pattern similar to the full date pattern, however, it also contains the code of the month. The following rules are used to expand these codes:
\begin{equation}
\begin{array}{l}
"\textit{«}01m\textit{ė}n\textit{»}" \rightarrow "sausio", \\
... \\
"\textit{«}12m\textit{ė}n\textit{»}" \rightarrow "gruod\textit{ž}io". \end{array} \nonumber
\end{equation}

E.g., \textit{Nuo 2013 01 04 \(\rightarrow\) Nuo «PG» 2013 01 04 \(\rightarrow\) Nuo 2013-ų metų «01mėn» «SG» 04 «FO» «SG» d. \(\rightarrow\) Nuo du tūkstančiai tryliktų metų sausio ketvirtos dienos}.

Short format dates without a preposition are expanded into the nominative case singular, e.g., \textit{2013 01 04 \(\rightarrow\) du tūkstančiai tryliktų metų sausio ketvirta diena}.
\begin{equation}
\begin{array}{l}
"\backslash b([12]\backslash d\backslash d\backslash d)[\ \textit{-}](0[1\textit{-}9]|10|11|12)[\ \textit{-}]([0\textit{-}3]\backslash d)\backslash b" \nonumber \\
\rightarrow "\$1\textit{-ų}\ met\textit{ų}\ \textit{«}\$2m\textit{ė}n\textit{»}\ \$3\ \textit{«}FO\textit{»}\ \textit{«}SN\textit{»}\ d.". \nonumber
\end{array} \nonumber
\end{equation}

\subsection{Years}
Although the year could be treated as an even shorter case of the date, there are several reasons why the year was made a separate class: 1) neither the month nor the day is specified next to the year, so the value itself needs to be checked more strictly to avoid confusion with a quantitative numeral or the like, 2) years in dates are always expanded into the genitive case, while written separately – usually into the instrumental case, 3) a range of years can be specified, however, this is not the case with dates.

After analyzing the data, it was decided to limit possible values of the year to the range 1500-2059. There are many examples from the \(16^{th}\) century, but no examples from the \(15^{th}\) century. This may be related to the fact that the first Lithuanian book - The Martynas Mažvydas Catechism - was printed in 1547 \cite{Sespl73}.

The year is expressed in ordinal numbers. If the number is followed by the word \textit{metai} with a certain inflection (\textit{metų, metus, metais}), the inflection of the word \textit{metai} denotes the inflection of the number of years. The inflection defined by the preposition (if any) is ignored, since it has a lower priority.

If the abbreviation \textit{m.} is used and there is a preposition, the preposition defines the inflexion of the year.

If the abbreviation \textit{m.} is used and there is no preposition to describe the inflection, the year is written in the instrumental case.
\begin{equation}
\begin{array}{l}
"(1[5\textit{-}9]\backslash d\backslash d|20[0\textit{-}5]\backslash d)\ (metus)" \rightarrow "\$1\textit{-}us\ \$2", \\
"(\textit{«}PG\textit{»})\ (1[5\textit{-}9]\backslash d\backslash d|20[0\textit{-}5]\backslash d)\ (m\backslash .)" \rightarrow "\$2\textit{-ų}\ \$1\ \$3", \\
"\backslash b(1[5\textit{-}9]\backslash d\backslash d|20[0\textit{-}5]\backslash d)\ (m\backslash .)" \rightarrow "\$1\textit{-}ais\ \textit{«}PI\textit{»}\ \$2". 
\end{array} \nonumber
\end{equation}

E.g., \textit{2001 metus \(\rightarrow\) 2001-us metus \(\rightarrow\) du tūkstančiai pirmus metus}.

E.g., \textit{iki 2021 m. \(\rightarrow\) iki «PG» 2001-ų «PG» m. \(\rightarrow\) iki du tūkstančiai pirmų metų}.

E.g., \textit{2001 m. \(\rightarrow\) 2001-ais «PI» m. \(\rightarrow\) du tūkstančiai pirmais metais}.

If the number belongs to the range 1500-2059 and the number is not followed by \textit{m.} or the word \textit{metai}, however, there is a punctuation mark, that number is treated as a year and is expanded into the pronominal form nominative case plural. This is justified in the vast majority of cases:
\begin{equation}
"\backslash b(1[5\textit{-}9]\backslash d\backslash d|20[0\textit{-}5]\backslash d)([,\backslash .;\backslash )]|\$)" \rightarrow "\$1\textit{-}ieji\$2". \nonumber
\end{equation}

E.g., \textit{Vilniaus universiteto leidykla, 2016. 105 p. \(\rightarrow\) Vilniaus universiteto leidykla, 2016-ieji. 105 p., „Šimtas kalbos mįslių“ (1970, 2001) \(\rightarrow\) „Šimtas kalbos mįslių“ (1970-ieji, 2001-ieji)}.

If there is no punctuation mark after the number, it may be the first term of the range of years. Ranges of years are another construction that makes it absolutely certain that the numbers indicate a year. If there is a preposition before the range of years, it defines the case (genitive or accusative), if there is no preposition – it is expanded into the instrumental case plural. E.g., \textit{nuo 2008/2009 m.m., apie 1665–1670 m., (1950 - 1997)}. The years can be separated by either a hyphen, a dash or a slash, with or without spaces. The grammatical form of the word to the right of the range of years was not analyzed and was not passed on to the left to determine the grammatical form of the numerals.
\begin{equation}
\begin{array}{l}
"\textit{«}PG\textit{»}\ ([12]\backslash d\backslash d\backslash d)\ ?[/-\ \textit{-}]\ ?([12]\backslash d\backslash d\backslash d)\backslash b" \rightarrow "\$1\textit{-ų}\ \$2\textit{-ų} \textit{«}PG\textit{»}",  \\
"\textit{«}PA\textit{»}\ ([12]\backslash d\backslash d\backslash d)\ ?[/-\ \textit{-}]\ ?([12]\backslash d\backslash d\backslash d)\backslash b" \rightarrow "\$1\textit{-}us\ \$2\textit{-}us \textit{«}PA\textit{»}", \\
"\backslash b([12]\backslash d\backslash d\backslash d)\ ?[/-\ \textit{-}]\ ?([12]\backslash d\backslash d\backslash d)\backslash b" \rightarrow "\$1\textit{-}ais\ \$2\textit{-}ais \textit{«}PI\textit{»}".
\end{array} \nonumber
\end{equation}

A special rule was written for the ranges of years in the following form: \textit{nuo 1981 iki 1986 metų}.
\begin{equation}
\begin{array}{l}
"(\textit{«}PG\textit{»})\ ([12]\backslash d\backslash d\backslash d)\ iki\ (\textit{«}PG\textit{»})\ ([12]\backslash d\backslash d\backslash d)\ (m\backslash .|met\textit{ų})" \\
\rightarrow "\$2\textit{-ų}\ iki\ \$4\textit{-ų}\ \$1\ \$5".
\end{array} \nonumber
\end{equation}

\subsection{Time}
Hours are written in one or two digits (i.e., numbers from 0 to 24) and are expanded using an ordinal number in the accusative case singular. When specifying the time, the minutes are always written in two digits (i.e., numbers from 00 to 59) and are expanded using a cardinal number, i.e., expansion does not depend on whether it is about time or about duration in minutes. Hours and minutes are separated by a period or a colon. If there are no minutes, nothing is written or 00 is written, and minutes are skipped when expanding. At the end of the time-defining construct there must be the abbreviation \textit{val.}, otherwise it will be confused with other semiotic classes in most cases, e.g., with the result of a basketball game. The following regex can be used to check that the hours fall within a possible range of values: $\backslash d|1\backslash d|2[0\textit{4}]$ and minutes – $[0\textit{-}5]\backslash d$. Of course, it is possible to limit yourself to simpler inspections. Examples of time: \textit{10.00 val., 13:15 val., 9 val.} 
\begin{equation}
\begin{array}{l}
"\backslash b(\backslash d|1\backslash d|2[0\textit{-}4])[:\backslash .]00\ val\backslash ." \rightarrow "\textit{«}SA\textit{»}\ \$1\ \textit{«}FO\textit{»} \textit{«}SA\textit{»}\ val.", \\
"\backslash b(\backslash d|1\backslash d|2[0\textit{-}4])[:\backslash .]([0\textit{-}5]\backslash d)\ val\backslash ." \rightarrow "\textit{«}SA\textit{»} \$1\ \textit{«}FO\textit{»}\ \textit{«}SA\textit{»}\ val.\ \$2\ min.", \\
"\backslash b(\hat{\ }\backslash d|[\hat{\ },\backslash d]\backslash d|1\backslash d|2[0\textit{-}4])\ val\backslash ." \rightarrow "\textit{«}SA\textit{»}\ \$1\ \textit{«}FO\textit{»}\ \textit{«}SA\textit{»}\ val."
\end{array} \nonumber
\end{equation}

The rules work as follows: \textit{13:15 val. \(\rightarrow\) «SA» 13 «FO» «SA» val. 15 min. \(\rightarrow\) tryliktą valandą penkiolika minučių}. The third rule is more complex because if only hours without minutes are specified, they should not be confused with quantitative numerals such as \textit{38,5 val. per savaitę (38.5 hours per week)}. Hours are usually measured to one decimal place.

The time can be preceded by a preposition that takes the genitive case, e.g., \textit{iki 13 val., nuo 10.30 val.} Prepositions that take the accusative case can be ignored because time is already expanded into the accusative case, e.g., \textit{apie 23.55 val.}
\begin{equation}
\begin{array}{l}
"\textit{«}SG\textit{»}\ (\backslash d|1\backslash d|2[0\textit{-}4])[:\backslash .]00\ val\backslash ." \rightarrow "\textit{«}SG\textit{»}\ \$1 \textit{«}FO\textit{»}\ \textit{«}SG\textit{»}\ val.", \\
"\textit{«}SG\textit{»}\ (\backslash d|1\backslash d|2[0\textit{-}4])[:\backslash .]([0\textit{-}5]\backslash d)\ val\backslash ." \\
\rightarrow "\textit{«}SG\textit{»}\ \$1\ \textit{«}FO\textit{»}\ \textit{«}SG\textit{»}\ val.\ \textit{«}PG\textit{»}\ \$2\ min.", \\
"\textit{«}SG\textit{»}\ (\backslash d|1\backslash d|2[0\textit{-}4])\ val\backslash ." \rightarrow "\textit{«}SG\textit{»}\ \$1\ \textit{«}FO\textit{»}\ \textit{«}SG\textit{»}\ val."
\end{array} \nonumber
\end{equation}

Time ranges can be expanded in several ways: \textit{9-15 val. \(\rightarrow\) devintą penkioliktą valandomis} or with prepositions \textit{nuo devintos valandos iki penkioliktos valandos}. However, if minutes are specified, only the second option is valid, e.g., \textit{9:15-20:30 \(\rightarrow\) nuo devintos valandos penkiolikos minučių iki dvidešimtos valandos trisdešimties minučių}, therefore, time ranges were expanded using a construction with prepositions.
\begin{equation}
\begin{array}{l}
"([12]?\backslash d)([:\backslash .][0\textit{-}5]\backslash d)?[-\textit{-}]([12]?\backslash d)([:\backslash .][0\textit{-}5]\backslash d)?\ val\backslash ." \\
\rightarrow "nuo\ \$1\$2\ val.\ iki\ \$3\$4\ val."
\end{array} \nonumber
\end{equation}

E.g., \textit{13.55–14.15 val. \(\rightarrow\) nuo 13.55 val. iki 14.15 val.} – and we already know how to expand such constructions.

A similar idea was tried with ranges of years, e.g., \textit{2011-2012 m. \(\rightarrow\) du tūkstančiai vienuoliktais dvyliktais metais} or \textit{nuo 2011 m. iki 2012 m.} However, ranges of years are very often used to describe two contiguous years, such as the school year, in which case only the first rather than second option is suitable so ranges of years are always expanded using the first construction.

If prepositions are already present, all you have to do is to insert the abbreviation \textit{val.} after the first component and we already have a familiar construction, e.g., \textit{nuo 9:00 iki 16:30 val. \(\rightarrow\) nuo 9:00 val. iki 16:30 val.}
\begin{equation}
\begin{array}{l}
"nuo\ ([12]?\backslash d)([:\backslash .][0\textit{-}5]\backslash d)?\ iki\ ([12]?\backslash d)([:\backslash .][0\textit{-}5]\backslash d)?\ val\backslash ." \nonumber \\
\rightarrow "nuo\ \$1\$2\ val.\ iki\ \$3\$4\ val." 
\end{array} \nonumber
\end{equation}

In addition, special rules were written to expand this type of time constructs: \textit{tarp 18.55 val. ir 21.17 val., tarp 3 ir 4 val. (between 3 and 4 o'clock)}.
\begin{equation}
\begin{array}{l}
"(\textit{«}S[GAI]\textit{»})\ (\backslash d\{1,2\}[:\backslash .]\backslash d\{2\}) (val\backslash .)\ (ir|arba)\ (\backslash d\{1,2\}[:\backslash .]\backslash d\{2\}) (val\backslash .)" \\
\rightarrow "\$1\ \$2\ \$3\ \$4\ \$1\ \$5\ \$6", \\
"(\textit{«}S[GAI]\textit{»})\ (\backslash d\{1,2\})\ (val\backslash .)\ (ir|arba)\ (\backslash d\{1,2\})\ (val\backslash .)" \\
\rightarrow "\$1\ \$2\ \$3\ \$4\ \$1\ \$5\ \$6".
\end{array} \nonumber
\end{equation}

\subsection{Phone numbers}
The format of telephone numbers in Lithuania is governed by the Order of the Director of the Communications Regulatory Authority of the Republic of Lithuania on the Approval of National and International Telephone Numbering Guidelines. 2005 December 23 No. 1V-1162 
(\url{https://www.e-tar.lt/portal/lt/legalAct/TAR.051C326E854A/TAIS\_300743}). The words \textit{Telefonas, Faksas, Mobilusis telefonas} or abbreviations of these words \textit{Tel., Faks., Mob.} should be written before the telephone number. The national telephone number consists of a destination code of 1, 2 or 3 digits and the corresponding 7, 6, or 5-digit network termination number. The international telephone number consists of the country code and the 8-digit national telephone number. If the national telephone number is given, it is preceded by the national prefix 8, which is enclosed in brackets together with the geographical destination code, e.g., \textit{Tel. (8 5) 123 4567}. If the international telephone number is given, it is preceded by the plus sign followed by the country code, such as \textit{Mob. + 370 123  45 678}. The numbers are grouped by means of a special method using spaces, the destination code and the end point number are separated by double spacing. Telephone numbers are not always written in accordance with the rules, so certain requirements can be ignored. E.g., \textit{tel. 123 45 67, Tel. (8 5) 123 4567, Tel. 8 612 34567, T.: +370 612 34567}. Expansion rules simply insert spaces between digits.
\begin{equation}
\begin{array}{l}
"(T|t)(\backslash .:|el\backslash .)\ (\backslash d)(\backslash d)(\backslash d)\ ?(\backslash d)\ ?(\backslash d)\ ?(\backslash d)\ ?(\backslash d)" \\
\rightarrow "\$1elefonas\ \$3\ \$4\ \$5\ \$6\ \$7\ \$8\ \$9", \\
"(T|t)(\backslash .:|el\backslash .)\ \backslash (?8[\ \textit{-}](\backslash d)\backslash )?[\ \textit{-}]?(\backslash d)(\backslash d)\ ?(\backslash d)\ ?(\backslash d)\ ?(\backslash d)\ ?(\backslash d)\ ?(\backslash d)" \\
\rightarrow "\$1elefonas\ 8\ \$3\ \$4\ \$5\ \$6\ \$7\ \$8\ \$9\ \$10", \\
"(T|t)(\backslash .:|el\backslash .)\ \backslash (?\backslash\textit{+}(\backslash d)(\backslash d)(\backslash d)\ ?(\backslash d)\backslash )?\ ?(\backslash d)(\backslash d)\ ?(\backslash d)\ ?(\backslash d)\ ?(\backslash d)\ ? \\ 
(\backslash d)\ ?(\backslash d)" \rightarrow "\$1elefonas\ plius\ \$3\ \$4\ \$5\ \$6\ \$7\ \$8\ \$9\ \$10\ \$11\ \$12\ \$13".
\end{array} \nonumber
\end{equation}

\subsection{Codes}
Until January 1, 2007, ISBNs were 10 digits, later they changed to 13 digits (\url{https://www.isbn.org/about_isbn_standard}). E.g., \textit{ISBN: 1909232424, ISBN 978-609-420-425-8}.
\begin{equation}
\begin{array}{l}
"ISBN:?\ (\backslash d)(\backslash d)(\backslash d)\textit{-}?(\backslash d)\textit{-}?(\backslash d)\textit{-}?(\backslash d)\textit{-}?(\backslash d)\textit{-}?(\backslash d)\textit{-}?(\backslash d)\textit{-}?(\backslash d)\textit{-}?(\backslash d)\textit{-}? \\
(\backslash d)\textit{-}?(\backslash d)" \rightarrow "I\textit{-}eS\textit{-}B\textit{ė}\textit{-}e\ \grave{ }\ N\ \$1\ \$2\ \$3\ \$4\ \$5\ \$6\ \$7\ \$8\ \$9\ \$10\ \$11\ \$12\ \$13", \\
"ISBN:? (\backslash d)(\backslash d)(\backslash d)\textit{-}?(\backslash d)\textit{-}?(\backslash d)\textit{-}?(\backslash d)\textit{-}?(\backslash d)\textit{-}?(\backslash d)\textit{-}?(\backslash d)\textit{-}?(\backslash d)" \\ 
\rightarrow "I\textit{-}eS\textit{-}B\textit{ė}\textit{-}e\ \grave{ }\ N\ \$1\ \$2\ \$3\ \$4\ \$5\ \$6\ \$7\ \$8\ \$9\ \$10".
\end{array} \nonumber
\end{equation}

Here hyphens are used to mark syllable boundaries. Besides the accent mark on the last syllable helps the synthesizer read the abbreviation correctly.

All five-digit numbers in the NAV and FLF datasets were postal codes, e.g., \textit{03123 Vilnius}. There were no postcodes in the DEL dataset, and the five-digit numbers represented cardinal numerals.
\begin{equation}
"(\backslash d)(\backslash d)(\backslash d)(\backslash d)(\backslash d)" \rightarrow "\$1\ \$2\ \$3\ \$4\ \$5". \nonumber
\end{equation}

A special rule for study program codes (dataset FLF) was also formulated, e.g., \textit{Rusų filologija (6121NX025)}.
\begin{equation}
"\backslash ((\backslash d)(\backslash d)(\backslash d)(\backslash d)NX(\backslash d)(\backslash d)(\backslash d)\backslash )" \rightarrow "(\$1\ \$2\ \$3\ \$4\ NX\ \$5\ \$6\ \$7)". \nonumber
\end{equation}

\subsection{URLs}
This class includes e-mail addresses and Internet addresses, e.g., \textit{el. p. vardas.pavardenis @cr.vu.lt \(\rightarrow\) elektroninis paštas vardas taškas pavardenis eta cė-e\(\ \grave{ }\)r taškas vė-u\(\ \grave{ }\) taškas el-tė\(\tilde{\ }\), http://www.prusistika.flf.vu.lt/zodynas/ apie/Pristatymas \(\rightarrow\) haš-tė-tė-pė\(\tilde{\ }\) vė-vė-vė\(\tilde{\ }\) taškas prusistika taškas ef-el-e\(\ \grave{ }\)f taškas vė-u\(\ \grave{ }\) taškas el-tė\(\tilde{\ }\) zodynas apie Pristatymas}. In this case, no attempt was made to create regular expressions that would identify urls, but simply a few rules that replaced certain elements specific to the url were created, e.g., deletes the slash (/) if it is not between numbers, replaces the period (.) with the word \textit{taškas (period)} if there are at least two letters to the left and right and no spaces.
\begin{equation}
\begin{array}{l}
"@" \rightarrow "\ eta\ ", \\
"://" \rightarrow "\ ", \\
"([\hat{\ }\backslash d\backslash s])/" \rightarrow "\$1\ ", \\
"/([\hat{\ }\backslash d\backslash s])" \rightarrow "\ \$1", \\
"([a\textit{-}z][a\textit{-}z])\backslash .([a\textit{-}z][a\textit{-}z])" \rightarrow "\$1\ ta\textit{š}kas\ \$2".
\end{array} \nonumber
\end{equation}

\subsection{Abbreviations read as words (ASWD)}
Many authors take the view that the non-standard word is the word that is not found in the dictionary of standard words. We, like \cite{Gerazov11}, will take the opposite view, i.e., that the non-standard word is the word that is found in the dictionary of non-standard words, all other words that we can read as ordinary words may not be considered non-standard words, i.e., they are not included in the class ASWD and they are not processed by the normalization algorithm. So, usually there is no need to do anything, however, a few things can be done:

1) Replace non-Lithuanian letters with Lithuanian ones, e.g., \textit{Czesławą Miłoszą \(\rightarrow\) Czeslavą Miloszą. Münsterio \(\rightarrow\) Miunsterio}.
\begin{equation}
\begin{array}{l}
"\textit{ł}" \rightarrow "l", \\
"\textit{ü}" \rightarrow "iu".
\end{array} \nonumber
\end{equation}

2) Remove the period at the end of the abbreviation so as not to confuse it with the end of the sentence; put the accent mark, e.g., \textit{prof. \(\rightarrow\) pro\(\ \grave{\ }\)f, BUS \(\rightarrow\) BU\(\ \grave{\ }\)S (Bendrosios universitetinės studijos, General university studies), ARKSI \(\rightarrow\) ARKSI\(\ \grave{\ }\) (Anglistikos, romanistikos ir klasikinių studijų institutas, Institute for English, Romance Studies and Classical Studies)}.
\begin{equation}
\begin{array}{l}
"\backslash b([pP])rof\backslash ." \rightarrow "\$1ro\ \grave{\ }f", \nonumber \\
"\backslash bBUS\backslash b" \rightarrow "BU\ \grave{\ }S", \\
"\backslash bARKSI\backslash b" \rightarrow "ARKSI\ \grave{\ }", 
\end{array} \nonumber
\end{equation}

3) Provide an orthographic transcription of some foreign words used in navigation that mean street, road, avenue, lane, etc.
\begin{equation}
\begin{array}{l}
"\backslash bUlica\backslash b" \rightarrow "Uly\tilde{\ }ca", \\
"\backslash bStrasse\backslash b" \rightarrow "\textit{Š}tra\tilde{\ }se", \\
"\backslash bStrada\backslash b" \rightarrow "Stra\tilde{\ }da", \\
"\backslash bAvenue\backslash b" \rightarrow "Aveniu\ \grave{\ }",
\end{array} \nonumber
\end{equation}

\subsection{Letter sequences (LSEQ)}
If a non-standard word is composed \textbf{entirely of consonants}, this fact can be easily detected and the word can be spelled without using a pronunciation dictionary. E.g., \textit{LSP \(\rightarrow\) eL-eS-Pė\(\tilde{\ }\) (Lietuvos studento pažymėjimas, Lithuanian student certificate), NPD \(\rightarrow\) eN-Pė-Dė\(\tilde{\ }\) (Neapmokestinamasis pajamų dydis, Exempt income)}. As has already been mentioned, regular expressions cannot expand abbreviations of any length, so 7 rules were written that process abbreviations of 1-7 letters in length, i.e., they insert \textit{«Spell»} tags. The following is a rule for a 3-letter abbreviation. Another important thing that we noticed is that words built as a sequence of letters always bear an accent on the last syllable, so a different tag (\textit{«SpellA»}) is inserted before the last letter.
\begin{equation}
\begin{array}{l}
"(\hat{\ }|[\backslash s\backslash (\backslash \{\backslash [\textit{„}\textit{-}])([b\textit{-}df\textit{-}hj\textit{-}np\textit{-}tv\textit{-}xz\textit{čšž}B\textit{-}DF\textit{-}HJ\textit{-}NP\textit{-}TV\textit{-}XZ\textit{ČŠŽ}]) \\
([b\textit{-}df\textit{-}hj\textit{-}np\textit{-}tv\textit{-}xz\textit{čšž}B\textit{-}DF\textit{-}HJ\textit{-}NP\textit{-}TV\textit{-}XZ\textit{ČŠŽ}]) \\
([b\textit{-}df\textit{-}hj\textit{-}np\textit{-}tv\textit{-}xz\textit{čšž}B\textit{-}DF\textit{-}HJ\textit{-}NP\textit{-}TV\textit{-}XZ\textit{ČŠŽ}]) \\
([\backslash s\backslash .,:;\textit{-}-\backslash\textit{+“}\backslash )\backslash \}\backslash ]]|\$)" \rightarrow "\$1\textit{«}Spell\textit{»}\$2\textit{«}Spell\textit{»}\$3\textit{«}SpellA\textit{»}\$4\$5".
\end{array} \nonumber
\end{equation}

Subsequently, the letters preceded by \textit{«Spell»} or \textit{«SpellA»} are replaced with letter names using the following rules:
\begin{equation}
\begin{array}{l}
"(\textit{«}Spell\textit{»})([bcdgptvz\textit{čž}BCDGPTVZ\textit{ČŽ}])" \rightarrow "\$2\textit{ė-}", \\
"(\textit{«}SpellA\textit{»})([bcdgptvz\textit{čž}BCDGPTVZ\textit{ČŽ}])" \rightarrow "\$2\textit{ė}\tilde{\ }\ ", \\
"(\textit{«}Spell\textit{»})([flmnrs\textit{š}FLMNRS\textit{Š}])" \rightarrow "e\$2\textit{-}", \\
"(\textit{«}SpellA\textit{»})([flmnrs\textit{š}FLMNRS\textit{Š}])" \rightarrow "e\ \grave{ }\ \$2".
\end{array} \nonumber
\end{equation}

For words that \textbf{contain vowels} and need to be spelled, such as \textit{VU SA (Vilniaus universiteto Studentų atstovybė, Vilnius University Student Representation), BKKI (Baltijos kalbų ir kultūrų institutas, Baltic Institute of Languages and Cultures)}, a list of regular expressions should be compiled, one for each abbreviation. If no regex is created for an abbreviation, it remains unexpanded and will be read as a word once it enters the synthesizer, probably completely incomprehensible. One of the solutions is to write the rules that spell out all words that are short (2-3 letters) and consist of only uppercase letters; however, this was not implemented in this work, and rules were written for all these abbreviations. Some abbreviations are taken from other languages, although it is often customary to spell them using the original letter names rather than Lithuanian ones, we will use the Lithuanian letter names as recommended in \cite{Grigas08}, e.g., \textit{IT \(\rightarrow\) I-Tė\(\tilde{\ }\)}.
\begin{equation}
\begin{array}{l}
"\backslash bVU\backslash b" \rightarrow "V\textit{ė}\textit{-}U\ \grave{ }\ ", \\
"\backslash bSA\backslash b" \rightarrow "eS\textit{-}A\tilde{\ }\ ", \\
"\backslash bBKKI\backslash b" \rightarrow "B\textit{ė}\textit{-}Ka\textit{-}Ka\textit{-}I\ \grave{ }\ ".
\end{array} \nonumber
\end{equation}

\subsection{Abbreviations to be expanded (EXPN)}
Lithuanian is a highly inflectional language. This fact can be a cause of many grammatical errors when trying to expand abbreviations into a full text.

Units of measure after numbers are usually expanded, e.g., \textit{16\%, 5 min., iki 1 proc., iki 18 °C}. The grammatical form of a unit of measure is usually determined by the tag conveyed by the number.

There are many well-established abbreviations that are always expanded in the same way without running the risk of making a mistake, e.g., \textit{Reg. Nr. \(\rightarrow\) Registracijos numeris (Registration number), Eil. Nr. \(\rightarrow\) Eilės numeris (Serial number), Nr. \(\rightarrow\) Numeris (Number), Tel. \(\rightarrow\) Telefonas (Phone), a. a. \(\rightarrow\) amžiną atilsį (rest in peace), el. p. \(\rightarrow\) elektroninis pastas (e-mail), angl. \(\rightarrow\) angliškai (in English), L. e. p. \(\rightarrow\) Laikinai einantis pareigas (Acting temporarily), pvz., \(\rightarrow\) pavyzdžiui (for example), Š. m. \(\rightarrow\) Šių metų (this year), t. y. \(\rightarrow\) tai yra (that is)}.
\begin{equation}
\begin{array}{l}
"\backslash bReg\backslash .\ ?Nr\backslash ." \rightarrow "Registracijos\ numeris", \\
"\backslash bEil\backslash .\ ?Nr\backslash ." \rightarrow "Eil\textit{ė}s\ numeris", \\
"\backslash bNr\backslash ." \rightarrow "Numeris ", \\
"\backslash b([tT]el)\backslash ." \rightarrow "\$1efonas", \\
"\backslash ba\backslash .\ ?a\backslash ." \rightarrow "am\textit{ž}in\textit{ą}\ atils\textit{į}", \\
"\backslash b([eE]l)\backslash .\ ?p\backslash ." \rightarrow "\$1ektroninis\ pa\textit{š}tas", \\
"\backslash b([aA]ngl)\backslash ." \rightarrow "\$1i\textit{š}kai", \\
"\backslash b([lL])\backslash .\ ?e\backslash .\ ?p\backslash ." \rightarrow "\$1aikinai\ einantis\ pareigas", \\
"\backslash b([pP])vz\backslash ." \rightarrow "\$1avyzd\textit{ž}iui", \\
 "\backslash bt\backslash .\ ?y\backslash ." \rightarrow "tai\ yra".
\end{array} \nonumber
\end{equation}

There are abbreviations that are used only at the end of a sentence, so a period is left after them, e.g., \textit{ir kt. \(\rightarrow\) ir kita. (etc.), ir pan. \(\rightarrow\) ir panašiai. (and the like), ir t. t. \(\rightarrow\) ir taip toliau. (and so on)}.
\begin{equation}
\begin{array}{l}
"\backslash bir\ kt\backslash ." \rightarrow "ir\ kita.", \\
"\backslash bir\ pan\backslash ." \rightarrow "ir\ pana\textit{š}iai.", \\
"\backslash bir\ t\backslash .\ ?t\backslash ." \rightarrow "ir\ taip\ toliau.".
\end{array} \nonumber
\end{equation}

There are abbreviations that we can read like a word, or try to expand and add an appropriate suffix, e.g., \textit{doc., prof., vyr.}, however, to identify the grammatical form correctly it is necessary to create many complex rules, running the risk of making mistakes in determining the grammatical form, or not identifying them at all, so it is more reliable to leave them as they are. E.g., \textit{vyr. redaktorės \(\rightarrow\) vyriausiosios redaktorės, vyr. redaktorė \(\rightarrow\) vyriausioji redaktorė, vyr. redaktorius \(\rightarrow\) vyriausiasis redaktorius (Chief Editor)}.
\begin{equation}
\begin{array}{l}
"\backslash b([vV]yr)\backslash .\ ?([a\textit{-}zA\textit{-}Z]\textit{+ė}s)" \rightarrow "\$1iausiosios\ \$2", \\
"\backslash b([vV]yr)\backslash .\ ?([a\textit{-}zA\textit{-}Z]\textit{+ė})" \rightarrow "\$1iausioji\ \$2", \\
"\backslash b([vV]yr)\backslash .\ ?([a\textit{-}zA\textit{-}Z]\textit{+}s)" \rightarrow "\$1iausiasis\ \$2".
\end{array} \nonumber
\end{equation}

In quite the same way, when talking about abbreviations without vowels, we are faced with the dilemma about whether to spell them or try to expand them using a set of complex rules. E.g., \textit{Šv. Jonų \(\rightarrow\) Šventų Jonų, šv. Kalėdų \(\rightarrow\) Šventų Kalėdų, Šv. Mišios \(\rightarrow\) Šventos Mišios, Šv. Onos \(\rightarrow\) Šventos Onos, Šv. Jeronimo \(\rightarrow\) Švento Jeronimo}. Similarly, the advice is not to expand.
\begin{equation}
\begin{array}{l}
"\backslash b([\textit{šŠ}]v)\backslash .\ ?([a\textit{-}zA\textit{-}Z\textit{ąčęėįšūž}]\textit{+ų})" \rightarrow "\$1ent\textit{ų}\ \$2", \\
"\backslash b([\textit{šŠ}]v)\backslash .\ ?([a\textit{-}zA\textit{-}Z\textit{ąčęėįšūž}]\textit{+}os)" \rightarrow "\$1entos\ \$2", \\
"\backslash b([\textit{šŠ}]v)\backslash .\ ?([A\textit{-}Z])" \rightarrow "\$1ento\ \$1".
\end{array} \nonumber
\end{equation}

\section{Experiments}\label{s7} 
There are lots of software implementations of regular expressions; we used the std::regex library (\url{https://en.cppreference.com/w/cpp/regex}) and C++ programming language. For text processing, we used the function std::regex\_replace, to which the text to be processed, the search template (left side of the rule) and the replacement template (right side of the rule) are passed on as parameters. All created rules were placed in the same file, without being divided into semiotic classes or the like. This made it possible not to duplicate the rules. An alternative way would be to group them into semiotic classes, which would require duplication of some rules, e.g., duplication of numeral expansion rules in other classes: NDATE, NYEAR, NTIME, etc. The rules are applied sequentially and therefore are sorted so that the rules with the widest context are applied first, e.g., long format dates with a range of days, then simple long format dates, dates without days specified, years, quantitative numerals. This way of storing the rules also has disadvantages: it is difficult to ensure that the application of one rule has no unexpected effect on another; it is difficult to find a place in the sequence of rules for a newly created rule; rules for handling a single item (such as date) are scattered across multiple locations, etc.

The data tables were sorted by NSW columns and then by Entries. This allowed the creation and grouping of expansion rules by NSW classes. Regex-based rules for expanding data table entries (see Table 3) were written manually.

It is noteworthy that rules were not created for all the entries. Rule development was stopped when it seemed that the created rule would only be applied to a specific NSW expansion, but it was unlikely to be used ever again. Of course, this is a rather subjective criterion. Number of rules and unexpanded entries are shown in Table~\ref{tbl-4}.

\begin{table}
\caption{Number of rules and unexpanded entries by data set.}\label{tbl-4}
\begin{tabular}{ l l l l }
\hline
& DEL & FLF & NAV \\
\hline
Total rules & 1590 & 1590 & 771 \\
Total entries (containing NSW) & 4057 & 8523 & 6091 \\
Unexpanded/incorrectly expanded entries & 625 (15,4\%) & 635 (7,5\%) & 7 (0,1\%) \\
\hline
\end{tabular}
\end{table}

The same set of regex rules was used for the DEL and FLF data sets and a different set of rules for the NAV data set because some rules were incompatible, i.e., different datasets required different NSW expansion, for example, in the NAV set, both \textit{m} and \textit{m.} had to be expanded into \textit{metrai (meters)}, although in other sets \textit{m.} was usually expanded into \textit{metai (years)}.

For the NAV set, it would be possible to write rules for all NSW, while there remain many unexpanded/incorrectly expanded NSWs in other two sets. Moreover, it is worth commenting on why there are so many rules. Redundant rules arose for the following reasons: if a rule was created for the nominative case, rules for the genitive, accusative and instrumental cases were created at the same time too. E.g., \textit{vienas-asis \(\rightarrow\) pirmasis, vienas-ojo \(\rightarrow\) pirmojo, vienas-ąjį \(\rightarrow\) pirmąjį, vienas-uoju \(\rightarrow\) pirmuoju}. If a rule was created for the digit 1, rules for other digits 2-9 were created at the same time, for example, if a rule \textit{1-asis \(\rightarrow\) pirmasis} was created, the rules \textit{2-asis \(\rightarrow\) antrasis, …, 9-asis \(\rightarrow\) devintasis} were also created, etc.

So far, all the rules have been created based on all available data. It is common to use two different sets: one data set is used to create rules and the other is used to test them. We solved this dilemma in a somewhat peculiar way: we split all the data into two approximately equal parts, from all the rules we selected those that were used to expand NSWs found in the first part (in this way we simulated the rulemaking based on the first part). Then we used the already selected rules to expand the NSWs found in the second part. Two experiments were conducted with the data split in two ways: I) the data were split by about half. II) each article was split approximately in half (in the case of NAV set, the data from each trip was split in half). The number of NSWs and the total number of words in each part of each experiment are shown in Table~\ref{tbl-5}. Each experiment was then repeated by swapping the rulemaking and testing sets, and the average percentage was calculated.

\begin{table}
\caption{Distribution of data in rulemaking and testing sets.}\label{tbl-5}
\begin{tabular}{ l l l l }
\hline
& DEL & FLF & NAV \\
\hline
NSWs / total words in $1^{st}$ group (I) & 2424 / 53070 & 4901 / 50646 & 5752 / 24993 \\
NSWs / total words in $2^{nd}$ group (I) & 2864 / 47056 & 5857 / 50652 & 4887 / 24530 \\
NSWs / total words in $1^{st}$ group (II) & 2562 / 50131 & 5249 / 50387 & 5073 / 24430 \\
NSWs / total words in $2^{nd}$ group (II) & 2726 / 49997 & 5515 / 50913 & 5566 / 25093 \\
\hline
\end{tabular}
\end{table}

The number of rules selected on the basis of the rulemaking set and their number used in the test set are shown in Table~\ref{tbl-6}. As we can see, in Experiment II, the percentage of the rules used in the test set is higher, which means that the rulemaking set matches the test set better. We can also see that this percentage depends on the data set, it is lowest in the DEL set and highest in the NAV set.

\begin{table}
\caption{Number of rules used for rulemaking/testing.}\label{tbl-6}
\begin{tabular}{ l l l l }
\hline
& DEL & FLF & NAV \\
\hline
Rules train in $1^{st}$/test in $2^{nd}$ group (I) & 461/314 (66\%) & 551/407 (72\%) & 175/110 (76\%) \\
Rules train in $2^{nd}$/test in $1^{st}$ group (I) & 475/307 & 567 / 403 & 115/111 \\
Rules train in $1^{st}$/test in $2^{nd}$ group (II) & 468/342 (71\%) & 586/486 (81\%) & 162/144 (89\%) \\
Rules train in $2^{nd}$/test in in $1^{st}$ group (II) & 495/341 & 605/476 & 161/142 \\
\hline
\end{tabular}
\end{table}

After the data had been split into two parts, rules were selected based on the first part and testing was performed using the second part. Similarly, rules were selected based on the second part and testing was performed using the first part. The expanded rows were listed in the right-hand column \textit{Generated expansion} of the data table (see Table~\ref{tbl-3}) and compared with the \textit{Correct expansion} in the penultimate column. For entries with only one NSW, the comparison was made automatically, for those with more than one NSW, the expansions of all NSWs were manually checked. If at least a part of the expanded NSW failed to match the correct expansion (e.g., the ending of at least one word did not match), it was considered to be an error. The results are shown in Table~\ref{tbl-7}. It shows the number of errors for each class, the percentage of errors among that class of NSW, and the percentage of errors among all NSWs.

\begin{table}
\caption{Results of experiment I.}\label{tbl-7}
\begin{tabular}{ l l l l l }
\hline
& Code & DEL & FLF & NAV \\
\hline
Letters & EXPN & 363 44.8\% (24.5\%) & 321 18.2\% (20.1\%) & 47 1.2\% (12.2\%) \\
& LSEQ & 137 13.3\% (9.3\%) & 197 6.9\% (12.3\%) & 45 10.0\% (11.7\%) \\
& ASWD & 22 1.8\% (1.5\%) & 95 4.6\% (5.9\%) & 29 1.8\% (7.5\%) \\
\hline
Numbers & NUM & 560 38.9\% (37.9\%) & 158 20.6\% (9.9\%) & 99 2.2\% (25.7\%) \\
& NORD & 223 91.8\% (15.1\%) & 414 62.9\% (25.9\%) & 35 100\% (9.1\%) \\
& NTEL & - & 13 19.7\% (0.8\%) & - \\
& NTIME & 17 48.6\% (1.1\%) & 46 9.2\% (2.9\%) & 5 100\% (1.3\%) \\
& NDATE & 35 34.3\% (2.4\%) & 137 16.1\% (8.6\%) & - \\
& NYEAR & 113 33.3\% (7.6\%) & 184 19.4\% (11.5\%) & - \\
& NCODE & - & 24 18.9\% (1.5\%)	& - \\
\hline
Other & URL & 8 21.1\% (0.5\%) & 8 5.6\% (0.5\%) & - \\
& NONE & 1 100\% (0.1\%) & 2 25.0\% (0.1\%) & 125 72.3\% (32.5\%) \\
\hline
\multicolumn{2}{ l }{Total non-standard words} & 1479 28.0\% (100\%) & 1599 14.9\% (100\%) & 385 3.6\% (100\%) \\
\hline
\end{tabular}
\end{table}

An exactly analogous experiment was carried out, the only difference being that each article was divided into two parts. Results are shown in Table~\ref{tbl-8}.

\begin{table}
\caption{Results of experiment II.}\label{tbl-8}
\begin{tabular}{ l l l l l }
\hline
& Code & DEL & FLF & NAV \\
\hline
Letters & EXPN & 248 30.5\% (19.9\%) & 241 13.7\% (21.6\%) & 11 0.3\% (11.6\%) \\
& LSEQ & 95 9.2\% (7.6\%) & 87 13.5\% (21.2\%) & 14 3.1\% (14.7\%) \\
& ASWD & 18 1.4\% (1.4\%) & 31 1.5\% (2.8\%) & 12 0.8\% (12.6\%) \\
\hline
Numbers & NUM & 557 38.7\% (44.7\%) & 140 18.3\% (12.5\%) & 41 0.9\% (43.2\%) \\
& NORD & 185 76.1\% (14.9\%) & 286 43.5\% (25.6\%) & 12 34.3\% (12.6\%) \\
& NTEL & - & 13 19.7\% (1.2\%) & - \\
& NTIME & 14 40.0\% (1.1\%) & 32 6.4\% (2.9\%) & - \\
& NDATE & 35 34.3\% (2.8\%) & 99 11.6\% (8.9\%) & - \\
& NYEAR & 85 25.1\% (6.8\%) & 155 16.4\% (13.9\%) & - \\
& NCODE & - & 23 18.1\% (2.1\%) & - \\
\hline
Other & URL & 7 18.4\% (0.6\%) & 9 6.3\% (0.8\%) & - \\
& NONE & 1 100\% (0.1\%) & 2 25.0\% (0.2\%) & 5 2.9\% (5.3\%) \\
\hline
\multicolumn{2}{ l }{Total non-standard words} & 1245 23.6\% (100\%) & 1118 10.4\% (100\%) & 95 0.9\% (100\%) \\
\hline
\end{tabular}
\end{table}

When comparing the total number of errors in Tables~\ref{tbl-7} and \ref{tbl-8}, we notice that there are fewer errors in Table~\ref{tbl-8}. This was expected, as the data for rule making and testing are more consistent when splitting articles. This shows how important it is to use data that cover as many different cases as possible when creating rules.

A comparison of Tables~\ref{tbl-7} and \ref{tbl-8} can also be used to determine which NSW classes have experienced the greatest reduction in errors. For the DEL dataset, the number of errors was significantly reduced for EXPN, LSEQ, NORD, NYEAR, and remained almost unchanged for NUM, NTIME, NDATE. In other words, it is enough to implement the expansion of numbers and dates once and for all, and it works regardless of the data. A larger number of errors in the experiment I was due to the fact that the rule-making and testing datasets contained more different abbreviations than needed to be expanded or spelled. The number of errors for the FLF dataset was also reduced in ASWD, NTIME and NDATE, with a slight change in NUM. For the NAV dataset, the reduction in the number of errors was very significant for almost all NSWs. In addition, the NAV data set stands out among other data sets by the abundance of the NONE class and the number of errors in it. This class includes various control characters that need to be removed. The problem is that different control characters were generated by navigation software during the trips to different countries.

\section{Analysis of errors}\label{s8} 
\subsection{Years}
By limiting the year values to 1500-2059, in most cases the year was identified correctly and was not confused with the numbers; however, for this reason several times years were not identified in the following examples: \textit{(310–383 m.), kuri datuojama nuo 3700 iki 3500 metų prieš Kristaus gimimą (which dates from 3700 to 3500 years before the birth of Christ)}. 

If the number was followed by the abbreviation \textit{m.} and there was no preposition, it was expanded into instrumental case plural, e.g., \textit{1984 m. akademikas įvertintas \(\rightarrow\) tūkstantis devyni šimtai aštuoniasdešimt ketvirtais metais akademikas įvertintas (In 1984 the academic was awarded)}, however, there were cases where the context required the use of the genitive case, e.g., \textit{1975 m. vadovėlis \(\rightarrow\) tūkstantis devyni šimtai septyniasdešimt penktų metų vadovėlis (a textbook from 1975)}. The application of this rule allowed 70 entries to be expanded correctly, though mistakes were made in 20 cases.

If the number belonged to the range 1500-2059 and neither the abbreviation \textit{m.} nor the word \textit{metų} was found after the number, and there was a punctuation mark, that number was treated as a year and was expanded into the nominative case plural pronominal form. Four examples were found where expansion in this way was necessary, however, punctuation was missing: \textit{2017 Nr. 8, 2014 (4), (2018 pavasaris), (1956 ir 1971 m.)}, and there was an example where the first two numbers were expanded in that way, though they needed to be expanded in the same way as the third number, i.e., into the instrumental case plural: \textit{Jos vykdytos ir 1644, 1680, 1684 m. \(\rightarrow\) Jos vykdytos ir tūkstantis šeši šimtai keturiasdešimt ketvirtais, tūkstantis šeši šimtai aštuoniasdešimtais, tūkstantis šeši šimtai aštuoniasdešimt ketvirtais metais (They were also carried out in 1644, 1680, 1684.)}.

When expanding the ranges of years, the verification of numerical values was less strict, limited to the range 1000-2999, but this release of restrictions caused no additional errors. Most of the year range errors were made due to the fact that the context required a different inflection. E.g., \textit{2017–2018 akademinių metų (of the 2017-2018 academic year)} was expanded into the instrumental rather than the genitive case. In addition, the ability to identify ranges of years in which the end of the range was indicated by two digits was not implemented, e.g., \textit{1953–57 studijavo (studied in 1953-57), 2011-12 m. m. rudens semestre (in the fall semester of the 2011-12 academic year)}, because in the latter example it was not clear whether 12 stood for a year or a month. A total of 53 errors related to ranges of years were made (out of 186).

\subsection{Dates}
Dates written in a long format, in the absence of a preposition, are expanded into the accusative case. Errors occur if the context requires another grammatical form, e.g., \textit{2014 m. kovo 14 d. protokolai: \(\rightarrow\) du tūkstančiai keturioliktų metų kovo keturioliktos dienos protokolai:}, i.e., the genitive case. The same applies to the dates without years, e.g., \textit{rugsėjo 1 d. programą \(\rightarrow\) rugsėjo pirmos dienos programą} (genitive is required), \textit{pastebėjo, kad kovo 19 d. jau užimta \(\rightarrow\) pastebėjo, kad kovo devyniolikta diena jau užimta} (nominative is required).

No correction of the date inflection according to the grammatical form of the word \textit{diena} was implemented, as in the example: \textit{2014 m. balandžio 25 diena}.

Dates containing several days connected by a conjunction were not implemented either: \textit{2013 m. lapkričio 26 arba 28 dienos., 2014 m. spalio 22 ir 29 d.} Analogous cases with dates without a year are as follows: \textit{Birželio 2 ir 5 d., rugsėjo 7 d.  ir arba 14 d.}

The range of dates below was expanded as if they were two separate dates: \textit{2018 m. balandžio 23 d.–gegužės 7 d. \(\rightarrow\) du tūkstančiai aštuonioliktų metų balandžio dvidešimt trečią dieną–gegužės septintą dieną}. Ideally, there should have been the following: \textit{du tūkstančiai aštuonioliktų metų balandžio dvidešimt trečią–gegužės septintą dienomis}. Although the resulting expansion is not ideal, it is understandable enough that no special rule was developed for such date ranges.

Other reasons why a suitable regex could not be found to expand long-format dates are the following:
\begin{enumerate}
    \item a slash instead of a dash or hyphen: \textit{1692 m. rugpjūčio 19 / 29 d.}
    
    \item too many spaces: \textit{1939 m. sausio 1\ \ \ d., Spalio\ \ \ 28–31 d.}
    
    \item month in capital letters: \textit{iki VASARIO 28 d.}
    
    \item missing abbreviation \textit{d.}: \textit{nuo sausio 11!}
\end{enumerate}

Dates written in a short format, in the absence of a preposition, are expanded into the nominative case. Errors occur if the context requires another grammatical form, e.g., \textit{(žiūrėta 2018-01-05) \(\rightarrow\) (žiūrėta du tūkstančiai aštuonioliktų metų sausio penktą dieną), 2014-06-19 9 val. \(\rightarrow\) du tūkstančiai keturioliktų metų birželio devynioliktą dieną devintą valandą}, i.e., the accusative case should be used instead of the nominative one.

There were cases where the date was written as a hybrid of a long and short form. E.g., \textit{2014 m. 06 16, 2014 m. 06 10-11 d}. It is obvious that such examples were not  processed correctly.

Expanding ranges of days or months was not implemented for the dates written in a short form: \textit{2015 08 16–09 02, 2014-06-19 – 20, 2014-06-25 d. 9:00 val. - 06-27 d. 15:00 val.}

There were dates where the year or both the year and the month were missing: \textit{06.19, 23 d. 10 val.}, as well as dates in a non-Lithuanian format: \textit{1/12/2004}.

\subsection{Time}
Errors in time processing occurred for the following reasons:
\begin{enumerate}
    \item \textit{8 val. per 5 darbo dienas (8 hours in 5 working days), iki 10 val. per dieną (up to 10 hours a day)} – here quantitative numerals were confused with time.
    \item \textit{6:00 PM} – time in a non-Lithuanian format.
    \item \textit{10.30-13.00, 9:00–17:30, (20:55)} – the abbreviation \textit{val.} was not given after the time.
    \item \textit{pusryčiaujate 9 arba 10 val.} – the processing of times connected by a conjunction was not implemented.
\end{enumerate}

\subsection{Ordinal numerals}
Several mistakes were made where additional information was added to the auditorium or classroom numbers, or the range of them was specified, e.g., \textit{314 A-B aud., 115A (SFS) aud., 108-109 kab.}

Most mistakes occurred when Arabic numerals had to be treated as ordinal ones, however, they were treated as cardinal numerals because the criteria are unclear. E.g., \textit{aprašo 15 p. \(\rightarrow\) aprašo penkioliktas punktas (fifteenth paragraph of the description)}, current wrong expansion – \textit{aprašo penkiolika puslapių (fifteen pages of description)}, \textit{eidama 94 m. \(\rightarrow\) eidama devyniasdešimt ketvirtus metus (in the 94th year) – eidama devyniasdešimt keturi metai (going 94 years), 9–12 klasių \(\rightarrow\) devintų dvyliktų klasių (9th to 12th classes) – devyni–dvylika klasių (9 to 12 classes), 1 priedas \(\rightarrow\) pirmas priedas (first appendix) – vienas priedas (one appendix)}.

Attempts were made to adapt Roman numerals to the inflection of the word following it. Errors occured if the word to be matched was separated by other words, e.g., \textit{I studijų pakopos (1st cycle of studies)} must be treated in the same way as \textit{I pakopos (1st cycle)}, and \textit{XII Tarptautinį baltistų kongresą (XII International Congress of Balts)} should be treated as \textit{XII kongresą (XII Congress)}.

Other errors in expanding Roman numerals are as follows:
\begin{enumerate}
    \item \textit{I. Antikos literatūros} – ordinal numeral \textit{I. (pirmas)} treated as the initial.
    
    \item \textit{XVIII a. Vakarų Europos autorių veikalai (18th c. Works by Western European authors)} – expanded into the nominative case instead of the genitive one, because the capital letter \textit{V} was treated as the beginning of the sentence.
    
    \item \textit{išlikusi IV a.} – must be expanded into \textit{išlikusi ketvirto amžiaus (surviving fourth-century)} rather than \textit{išlikusi ketvirtas aukštas (surviving fourth-floor)}.
    
    \item \textit{nuo XIII a. (since the 13th century)} – for Roman numerals the alignment with the preposition was not implemented.
    
    \item \textit{XVIII a. susiklosčiusios istorinės aplinkybės (historical circumstances in the eighteenth century)} – the locative case should be used instead of the genitive one. Unclear criteria for determining the necessary grammatical form.
    
    \item \textit{XX–XXI a. pradžioje, I–II dalis, XIX ir XX amžiaus} – ranges and enumerations of Roman numerals were not implemented.
    
    \item \textit{Karolis XVI} – the pronominal form should be used \textit{Karolis šešioliktasis}. Unclear criteria for determining whether to expand it into the pronominal form.
\end{enumerate}

\subsection{Cardinal numerals}
It is quite common for a number without a preposition to be expanded into another (non-nominative) case, but the form of another case coincides with the nominative one, so that the correct result is still obtained. E.g., \textit{viršija 80 dolerių už barelį, \(\rightarrow\) viršija aštuoniasdešimt dolerių už barelį (exceeds eighty dollars per barrel)} – the accusative case should be used, however, but fortunately it coincides with the nominative one.

Some incorrectly processed examples and causes of errors:
\begin{enumerate}
    \item \textit{2 akademinės val.} – the unit of measure separated from the number by another word, the feminine form had to be used.
    
    \item \textit{per beveik 100 pozicijų} – the preposition separated from the number by another word, the accusative form had to be used.
    
    \item \textit{netekus 15 proc. biudžeto lėšų arba 300 tūkst. eurų} – the context requires the genitive case.
    
    \item \textit{už balsavus 35} – the context requires the dative case.
    
    \item \textit{sumokėjo 55 Eur} – the context requires the accusative case.
    
    \item \textit{padidino importo apimtį 59 \%} – the context requires the instrumental case.
    
    \item \textit{1 nemokama vieta} (nominative case), \textit{7 tomų} (gen.), \textit{4 padaliniams} (dat.), \textit{5 kandidatus} (acc.), \textit{4 balais} (instrumental), \textit{7 centruose} (locative) – the grammatical alignment with the word from the right was not implemented.
    
    \item \textit{13 638,47 EUR} – space between numbers.
    
    \item \textit{13 / 25 \(\rightarrow\) trylika iš dvidešimt penkių (13 out of 25)} – processing of such a construction was not implemented.
\end{enumerate}

\subsection{Phone numbers, Codes, URLs}
Phone numbers are easy to confuse with all sorts of numbers, so it was necessary for the phone number to be preceded by the word \textit{telefonas (phone)} or its abbreviation \textit{tel.} or \textit{T.} If it is not found, the number was not treated as a phone number and was expanded incorrectly. E.g., \textit{(8 5) 268 9999, Mob. +370 679 99999}.

Only codes for which regular expressions were written were expanded correctly. Unknown format codes, e.g., \textit{Nr. 09.3.3-LMT-K-712, 63543-LA-1-2014-1-LT-E4AKA1-ECHE}, were expanded incorrectly. It is hard to expect to be able to write regular expressions for all codes.

Only simple Internet addresses are expanded correctly. If a meaningless sequence of characters is found in the address, e.g., \textit{https://www.youtube.com/ watch?v=sqvGlmdNoSw}, it is likely to be expanded incorrectly.

\subsection{Expansion of abbreviations}
When expanding units of measure, errors can occur if, for example, the grammatical form of a number is defined incorrectly, then the error is passed on to the unit of measure. E.g., \textit{sumokėjo 55 Eur. (paid 55 Euros)} the accusative case should be used instead of the nominative one.

If you try to expand the abbreviation of class ASWD or LSEQ, errors can occur if the abbreviation is followed by an unknown word or a new grammatical form. E.g., the aforementioned abbreviation \textit{Šv. (Saint)} was successfully expanded 40 times, but a mistake was made when the word after the abbreviation was accented or divided into syllables: \textit{Šv. Jõnų, Šv. Ka-lė-dos}.

Another problem with abbreviations without vowels is that some of them can have a lot of meanings. E.g., abbreviations \textit{g., k., p.}:

g.: \textit{g. 1925-01-04 \(\rightarrow\) gimęs/gimusi (born) 1925-01-04, Universiteto g. 5 \(\rightarrow\) Universiteto gatvė (street) 5, po 200 g. \(\rightarrow\) po 200 gramų (grams)}.

k.: \textit{vokiečių k. \(\rightarrow\) vokiečių kalba (language), Malkavos k. \(\rightarrow\) Malkavos kaimas (willage), 242 k. \(\rightarrow\) 242 kabinetas (classroom), II k. magistrantė \(\rightarrow\) II kurso (course) magistrantė}.

p.: \textit{aprašo 15 p. \(\rightarrow\) aprašo penkioliktas punktas (paragraph), L. e. p. VU rektorius \(\rightarrow\) Laikinai einantis pareigas (Acting) VU rektorius, el. p. \(\rightarrow\) elektroninis pastas (mail), p. Petr Vavra \(\rightarrow\) ponas (Mister) Petr Vavra, žr. p. 25–31 \(\rightarrow\) žiūrėti 25–31 puslapius (pages)}.

\section{Conclusions}\label{s9} 
The present work examined the expanding of non-standard words in Lithuanian texts. The taxonomy of semiotic classes adapted to the Lithuanian language was presented in the work. Sets of rules were created based on regular expressions for NSW identification and expansion. Experiments were performed with three completely different data sets. Having divided the data in half by articles and using one half for rulemaking, the other half for testing, the following error rates were obtained in different sets: 23.6\% on the news portal, 10.4\% on the university website and 0.9\% on the car navigation generated texts. Most of the errors on the news portal were cardinal numerals for which it was impossible to identify their grammatical form correctly, and on the university website there were ordinal numerals that did not have any signs to distinguish them from cardinal ones. Another significant source of errors was the expansion of abbreviations when the grammatical form of the expanded word was incorrectly determined. It is recommended to expand only the units of measure following the number, and in other cases not to expand the abbreviations but to either read them as a word or spell them.

\bibliographystyle{plainurl}  
\bibliography{regex_PK_Arxiv}  

\end{document}